\documentclass[a4paper]{itor}

\usepackage[utf8]{inputenc}
\usepackage[T1]{fontenc}
\usepackage{amsmath,amssymb,amsthm}
\usepackage[cal=euler]{mathalfa}
\usepackage{mathptmx}  
\usepackage{booktabs}
\usepackage{array}
\usepackage{multirow}
\usepackage{algorithm}
\usepackage{algpseudocode}
\usepackage{hyperref}
\usepackage[numbers,sort&compress]{natbib} 
\usepackage{setspace}
\usepackage{xcolor}
\usepackage{tcolorbox}

\newtheorem{definition}{Definition}
\newtheorem{remark}{Remark}

\title{GenOR-Twin: A Semantic Middleware for Integrating Operational Discourse with Mathematical Optimization}

\author[Monemi et al.]{Rahimeh Neamatian Monemi, Shahin Gelareh,  Lubin Cui, Nelson Maculan}

\affil{Sharkey Predictim Globe}
\affil{D\'epartement R\&T, IUT de B\'ethune, Universit\'e d'Artois, F-62000 B\'ethune, France}
\affil{Federal University of Rio de Janeiro, COPPE-PESC, P.O. Box 68511, Rio de Janeiro, RJ 21941-972, Brazil}
\affil{School of Mathematics and Statistics, Henan Normal University, Xinxiang 453007, Henan, China}
\email{shahin.gelareh@univ-artois.fr}

\begin{document}
\begin{abstract}
We introduce GenOR-Twin, a neuro-symbolic framework that bridges the translation gap between unstructured operational logs and rigorous mathematical optimization. Our architecture uniquely positions Large Language Models as semantic translators rather than direct solvers, ensuring that the system retains the feasibility guarantees of exact combinatorial methods. { \color{blue}We design a dynamic constraint injection mechanism (the runtime translation of qualitative disruption events into formal mathematical constraints) that allows the system to structurally modify the optimization problem's feasibility region in real-time based on qualitative human inputs. The resulting bidirectional coupling---where operational observations update the virtual model state and optimized decisions are reflected back into the Knowledge Graph---satisfies the synchronization requirement of a proper Digital Twin. The framework features an adaptive decision policy} that automatically selects between low-complexity schedule repair and full re-optimization by analyzing the available system slack. Finally, we demonstrate the generalization of this approach across six distinct optimization domains, {\color{blue}turning static models into resilient systems that adapt to the operational uncertainty and variability of real-world environments.}

\end{abstract}

\maketitle




\section{Introduction}

The field of Operations Research has long provided the mathematical backbone for industrial efficiency. From the early days of linear programming in the 1940s to modern mixed-integer solvers capable of handling millions of variables, OR methods have saved organizations trillions of dollars by optimizing supply chains, production schedules, and logistics networks. Yet despite these advances, a fundamental disconnect persists between the theoretical elegance of optimization models and the messy reality of day-to-day operations.

Let us consider a typical manufacturing floor. A carefully optimized production schedule assumes that machines will operate according to specification, that raw materials will arrive on time, and that workers will be available as planned. In practice, none of these assumptions hold perfectly. A machine operator notices a subtle vibration and notes it in a logbook. A supplier sends an email warning of potential delays. A maintenance technician mentions during a shift handover that a hydraulic seal looks worn. These signals, rich with actionable information, exist entirely outside the mathematical models that govern production decisions.

This gap between qualitative operational intelligence and quantitative optimization represents what we term the ``Translation Gap.'' Traditional OR solvers require structured inputs: vectors of processing times, matrices of resource capacities, and precisely defined constraint inequalities. Real-world disruptions, however, often arrive as unstructured text, verbal communications, or implicit patterns that human operators recognize intuitively but cannot easily formalize for a computer system.

The emergence of Digital Twins has partially addressed this challenge by creating virtual replicas of physical systems that continuously ingest sensor data. Yet even sophisticated Digital Twins primarily serve descriptive and predictive functions. They can tell us what is happening and forecast what might happen, but they rarely prescribe complex corrective actions that respect the mathematical rigor required for true optimization.

Meanwhile, Generative Artificial Intelligence, particularly Large Language Models, has demonstrated remarkable capabilities in understanding and generating natural language. These models can parse complex documents, extract relevant entities, and reason about relationships in ways that seemed impossible just a few years ago. However, applying LLMs directly to optimization problems is fraught with difficulty. Neural networks are prone to arithmetic errors and can ``hallucinate'' solutions that violate hard constraints, making them unsuitable as standalone solvers for combinatorial optimization.

{\color{blue}
Recent work has explored adjacent positions in this design space. \citet{jackson2024nlsim} demonstrated that natural language can drive simulation-based decision models for operations management; however, their approach targets stochastic simulation rather than combinatorial optimization with hard feasibility guarantees. \citet{li2025generative} applied generative AI to process planning in manufacturing systems but without symbolic validation of constraint consistency. \citet{li2025cirp} introduced large manufacturing decision models for human-centric settings, yet without an automated constraint injection mechanism connecting unstructured discourse to solver inputs. None of these approaches simultaneously provide (a) automated real-time constraint injection from unstructured logs, (b) symbolic validation of feasibility, and (c) an adaptive decision rule for selecting between heuristic repair and full re-optimization---the triple capability that defines GenOR-Twin's contribution.
}

{\color{blue}This paper introduces GenOR-Twin, a semantic middleware designed to bridge these worlds by positioning Large Language Models as robust \textit{semantic translators} rather than arithmetic solvers. Throughout this paper, ``semantic'' is used in the computational linguistics sense: a semantic translation is a meaning-preserving mapping from a natural language expression to its formal symbolic representation. For example, ``Machine~3 is overheating'' is semantically translated into the mathematical constraint $S_{i,3} \geq t_{now} + 120$ for all pending operations on M3---preserving operational intent (resource unavailability) rather than merely matching syntactic patterns. Unlike Digital Twin architectures that primarily ingest physical data in a single direction, GenOR-Twin establishes a bidirectional coupling: operational observations drive constraint injection into the optimization model (physical-to-virtual direction), while optimized decisions feed back into the Knowledge Graph to update the virtual state (virtual-to-physical direction). In our architecture, an LLM-based agent acts as a decoder}
that perceives qualitative operational data and converts it into structured mathematical constraints. {\color{blue}The framework automates the translation process by integrating the LLM into the data-ingestion loop as a semantic parser that feeds symbolic solvers; this operation is formalized as \textit{dynamic constraint injection} (Definition~\ref{def:constraint_injection}, Section~\ref{sec:formulation}).} Crucially, we utilize an \textit{Adaptive Smart Scheduler} that decided whether to apply low-complexity schedule repair or full re-optimization based on available system slack. This approach ensures that we maintain the mathematical rigor of classical optimization while gaining the linguistic flexibility of neural networks.

Our contributions are as follows: (1) we provide a practical engineering framework for the automated injection of dynamic constraints from natural language logs, (2) we describe a robust decision rule for choosing between schedule repair and re-optimization based on system state, and (3) we evaluate the cross-domain utility of this approach across multiple optimization domains. While the mathematical solvability remains the domain of established OR methods, the GenOR-Twin framework reduces the latency and human effort in the operation-to-model loop.

The remainder of this paper is organized as follows. In Section~\ref{sec:literature}, we review the relevant literature on Digital Twins, machine learning for combinatorial optimization, and reactive scheduling methods. In Section~\ref{sec:formulation}, we present the formal mathematical foundations for the six optimization problems addressed in this study. In Section~\ref{sec:methodology}, we describe the GenOR-Twin methodology in detail. Subsequent sections present computational experiments and managerial implications.

\section{Literature Review}\label{sec:literature}

The integration of artificial intelligence with operations research has attracted significant attention in recent years, driven by advances in machine learning and the increasing availability of operational data. This section reviews three streams of literature that inform our work: Cognitive Digital Twins, AI-augmented optimization, and dynamic scheduling under uncertainty.

\subsection{Digital Twins and Cognitive Systems}

The concept of the Digital Twin originated in manufacturing as a virtual representation of physical assets for simulation and monitoring purposes \citep{grieves2014}. Early implementations focused on lifecycle management and predictive maintenance, using sensor data to anticipate equipment failures. \citet{rasheed2020} provided a comprehensive survey of Digital Twin values, challenges, and enablers, identifying data integration and model fidelity as key concerns.

More recently, researchers have extended the Digital Twin paradigm toward ``Cognitive'' systems that incorporate reasoning capabilities \citep{zheng2021cognitive, eirinakis2020cognitive}. A comprehensive review by \citet{zhuang2024review} reinforces the transition toward intelligent shop-floor management through DT-assisted scheduling and production process control. However, most existing cognitive architectures rely on reinforcement learning or rule-based systems \citep{li2022reinforcement}, which require extensive training data or expert knowledge to encode. Our work differs by leveraging pre-trained language models that can interpret diverse operational scenarios without task-specific training.

{\color{blue}
In parallel, \citet{ivanov2026agentic} introduced Agentic Digital Twins that bridge model-based and AI-driven decision-making for supply chain management, and \citet{ivanov2026omega} demonstrated Digital Twin design and implementation at industrial scale---both works reinforcing the need for frameworks that couple AI reasoning with operational decision support under real-time constraints.
}

\subsection{Machine Learning for Combinatorial Optimization}

The intersection of machine learning and combinatorial optimization represents a rapidly growing research area. \citet{bengio2021} authored an influential survey categorizing approaches into end-to-end learning, learning to configure algorithms, and hybrid methods. More recently, the emergence of Large Language Models has opened a new frontier: \textit{optimization from natural language}. Projects like \textbf{OptiMUS} \citep{zelikman2024} and work by \citet{wu2023} have demonstrated that LLMs can generate mathematical formulations from natural language problem descriptions.\

{\color{blue}
While OptiMUS \citep{zelikman2024} and LLM4Opt position the LLM as a \textit{formulation generator}---producing full mathematical programs from natural language, including objective functions and decision variable definitions---GenOR-Twin takes a fundamentally different architectural stance. Here, the LLM is restricted to \textit{constraint-level semantic extraction}, producing structured JSON patches that are validated by a symbolic gatekeeper before reaching the solver. This design preserves the feasibility guarantees and optimality certificates of exact combinatorial methods, at the cost of requiring pre-defined constraint template schemas. Critically, GenOR-Twin never exposes the objective function or decision variables to the neural component, eliminating the risk of arithmetic hallucination at the formulation level.
}

However, using LLMs as direct solvers (e.g., through Chain-of-Thought reasoning for the Traveling Salesman Problem (TSP)) remains limited by their arithmetic instability \citep{yang2023large, valmeekam2023planning}.
Our framework differs from end-to-end neural solvers by positioning the LLM strictly as a \textit{semantic interface}—a neuro-symbolic bridge that converts qualitative operational intelligence into structured constraints for exact solvers, similar to the OptiGuide approach for supply chains \citep{li2023large}.

{\color{blue}
At the intersection of generative AI and operations management, \citet{simchilevi2025} argue for democratizing optimization through generative AI, positioning LLMs as accessibility layers that reduce the expertise barrier for exact solvers---a perspective closely aligned with GenOR-Twin's architectural stance. \citet{jackson2024genai} survey generative AI applications across supply chain and OR domains, identifying semantic translation from operational discourse to solver inputs as an open research challenge that the present work addresses.
}

\subsection{Knowledge-Centric Digital Twins}

The evolution of Digital Twins toward ``Cognitive'' systems has increasingly relied on Knowledge Graphs and semantic technologies to move beyond simple descriptive monitoring toward active reasoning \citep{zheng2021cognitive, lu2020digital}. \citet{rasheed2020} and \citet{negri2017review} identify data integration as a key challenge in industrial informatics, particularly when bridging the gap between low-level telemetry (Supervisory Control and Data Acquisition (SCADA) / Manufacturing Execution System (MES) data) and the high-level operational knowledge typically held by human experts. While rule-based event handlers have existed for decades, {\color{blue}they lack the flexibility to interpret the high-variability, unstructured communications common in manufacturing handovers}, where vital disruption signals are often buried in unstructured shift logs or informal alerts. \citet{zhuang2021digital} emphasize that resilient production management requires frameworks capable of dynamically adapting to such shop-floor variations. GenOR-Twin addresses this need by utilizing the emergent reasoning of Large Language Models to automate the populating and updating of Knowledge Graphs from unstructured logs. This transformation allows the Digital Twin to act as a more sophisticated semantic sensor, grounding human observations into the formal symbolic structure required for mathematical optimization.

\section{Problem Scope and Primary Formalization}\label{sec:formulation}

GenOR-Twin is designed as a domain-agnostic framework capable of supporting a wide range of combinatorial optimization problems. To demonstrate this flexibility, we evaluate the system across six distinct domains: Job Shop Scheduling (JSSP \cite{jackson1956}), Vehicle Routing (VRP \cite{toth2002}), Project Scheduling (RCPSP \cite{kolisch1997}), Nurse Scheduling (NSP \cite{burke2004}), Bin Packing (BPP \cite{martello1990lower}), and Maximum Flow \cite{edmonds1972}. While we maintain a full mathematical specification for each in Appendix~A, this section focuses on the formalization of our primary industrial use case, the Job Shop Scheduling Problem, followed by the general dynamic constraint model.

\subsection{Domain Taxonomy and Evaluation Scope}

The selection of these six problems is not arbitrary; they represent fundamentally different mathematical structures and industrial contexts, allowing us to stress-test the neuro-symbolic bridge's ability to map varied semantic inputs to diverse symbolic constraints. We classify these problems into four distinct categories as shown in Table~\ref{tab:taxonomy}.

\begin{table}[ht]
\centering
\caption{Cross-Domain Evaluation Suite and Constraint Mapping}
\label{tab:taxonomy}
\small
\begin{tabular}{llll}
\toprule
\textbf{Category} & \textbf{Problem} & \textbf{Primary Resource} & \textbf{Example Disruption} \\ 
\midrule
\textit{Discrete Scheduling} & JSSP & Machines ($M_i$) & "Spindle motor failure on M2" \\
 & RCPSP & Renewable Resources ($R_k$) & "Overlapping site maintenance" \\
 & NSP & Human Personnel ($N_i$) & "Nurse 4 has personal emergency" \\
\midrule
\textit{Spatial Routing} & CVRP & Vehicles ($K_j$) & "Truck 10 stuck in heavy traffic" \\ 
\midrule
\textit{Spatio-Temporal Packing} & BPP & Bin Capacity ($V$) & "Bin 5 safety seal damaged" \\
\midrule
\textit{Network Flow} & Max Flow & Edge Capacity ($u_{ij}$) & "Pipe segment 4-7 leakage" \\ 
\bottomrule
\end{tabular}
\end{table}

This taxonomy ensures that GenOR-Twin's \textit{Translator Agent} must reason about discrete availability (JSSP/NSP), resource intensity (RCPSP), spatial constraints (CVRP), volumetric capacity (BPP), and throughput constraints (Max Flow).

\subsection{Job Shop Scheduling Problem (JSSP)}

The Job Shop Scheduling Problem is one of the most studied problems in combinatorial optimization, introduced by \citet{jackson1956} and extensively analyzed by \citet{muth1963}.

\begin{definition}[JSSP Instance]\label{def:jssp_instance}
A Job Shop Scheduling Problem instance is formally defined as a tuple $\Pi = (\mathcal{J}, \mathcal{M}, \mathcal{O}, p, d, w)$, where $\mathcal{J} = \{1, \ldots, n\}$ represents the set of jobs to be scheduled and $\mathcal{M} = \{1, \ldots, m\}$ denotes the discrete set of available machines. {\color{blue}The set $\mathcal{O}$ comprises all operations $O_{j,k}$ associated with job $j \in \mathcal{J}$, where $k \in \{1,\ldots,|\mathcal{O}_j|\}$ is the operation index within job $j$ (i.e., the $k$-th processing step of job $j$), each requiring processing on a specific machine.} The function $p: \mathcal{O} \rightarrow \mathbb{Z}^+$ maps each operation to its deterministic processing time, while $d$ and $w$ assign a strict due date and priority weight to each job, respectively.
\end{definition}

\begin{definition}[Objective Function]\label{def:jssp_obj}
Minimize Total Weighted Tardiness:
\begin{equation}
\min Z = \sum_{j \in \mathcal{J}} w_j \cdot T_j \quad \text{where} \quad T_j = \max(0, C_j - d_j)
\label{eq:jssp_obj}
\end{equation}
\end{definition}

The JSSP is well-known to be NP-hard \citep{garey1979}.


\subsection{The Dynamic Constraint Injection Model}

The central challenge addressed by GenOR-Twin is the transition from the probabilistic world of natural language observations to the deterministic world of symbolic optimization. We formalize this bridge through the concept of \textit{dynamic constraint injection}.

\begin{definition}[Disruption Event]\label{def:disruption}
A disruption $\xi$ is a triple $(R, W, \delta)$ where $R \subseteq \mathcal{R}$ is the set of affected resources (e.g., specific machines or personnel), $W = [t_s, t_e]$ is the time window of unavailability, and $\delta$ is the semantic descriptor containing unstructured text (e.g., "The motor is overheating and needs 2 hours of cooling").
\end{definition}

\begin{definition}[Neuro-Symbolic Constraint Injection]\label{def:constraint_injection}
Given a disruption $\xi$, the neuro-symbolic translation function $\Phi$ acts as a mapping $\Phi: \Delta \rightarrow \mathcal{C}$, where $\Delta$ is the space of unstructured observations and $\mathcal{C}$ is the set of valid symbolic constraints for the underlying problem instance. For JSSP, this produces:
\begin{equation}
\Phi(\xi) = \{C_r^W : r \in R, C_r^W \text{ forbids resource } r \text{ during } W\}
\label{eq:phi}
\end{equation}
This injection modifies the feasible space of the optimization problem $X_{feasible}$ in real-time, necessitating an immediate re-computation of the optimal policy $\pi$.
\end{definition}

\section{Mathematical Formulation and System Context}\label{sec:or_contribution}

{\color{blue}\subsection{Stochastic Extension of the JSSP Objective Function: \eqref{eq:stochastic_obj} vs.\ \eqref{eq:jssp_obj}}\label{sec:smip}}

The proposed framework addresses the dynamic scheduling problem through a \textit{Single-Level Stochastic Mixed-Integer Program (SMIP)} formulation. Let $\xi$ be a random variable representing the semantic uncertainty of the disruption (e.g., duration $d(\xi)$, affected machine $m(\xi)$).

\begin{definition}[Stochastic Objective Function]\label{def:stochastic_obj}
We minimize the expected total weighted tardiness:
\begin{equation}
\min_{\mathbf{x}} \quad \mathbb{E}_{\xi} \left[ \sum_{j \in \mathcal{J}} w_j T_j(\mathbf{x}, \xi) \right]
\label{eq:stochastic_obj}
\end{equation}
subject to standard precedence and capacity constraints.
\end{definition}
{\color{blue}
\begin{remark}
This formulation extends Definition~\ref{def:jssp_obj} (the deterministic JSSP objective, \eqref{eq:jssp_obj}) by introducing the expectation operator $\mathbb{E}_{\xi}[\cdot]$ over the random disruption variable $\xi$, which encodes semantic uncertainty in disruption duration $d(\xi)$ and affected machine $m(\xi)$. When $\xi$ is deterministic---i.e., when the LLM provides a point estimate $\hat{\xi}$ with confidence $\theta > \tau$---\eqref{eq:stochastic_obj} reduces exactly to \eqref{eq:jssp_obj}. In practice, the SmartScheduler solves the deterministic version after the LLM resolves $\hat{\xi}$; \eqref{eq:stochastic_obj} is the theoretical framing that motivates the confidence-based policy $\Pi(D, S, \theta)$.
\end{remark}
}

\begin{definition}[Second-Stage Recourse]\label{def:recourse}
Let $\hat{\xi}$ be the realization inferred by the agent with confidence $\theta = \text{conf}(\hat{\xi})$. The recourse action $\mathbf{y}(\hat{\xi})$ minimizes the cost of deviation:
\begin{equation}
Q(\mathbf{x}, \hat{\xi}) = \min_{\mathbf{y}} \left( \alpha \cdot \text{Tardiness}(\mathbf{y}) + (1-\theta) \cdot \text{RiskPenalty} \right)
\label{eq:recourse}
\end{equation}
\end{definition}

Our \textit{SmartScheduler} (Section 5) acts as a computationally efficient approximation of this SMIP, using the confidence score $\theta$ to toggle between cheap recourse (Right-Shift) and expensive recourse (Re-optimization).

\subsection{Methodological Positioning}
Traditional OR models assume that exogenous information enters the system in a structured, quantitative form, such as parameters, coefficients, or pre-defined constraints. In practice, however, a substantial portion of operational intelligence arrives as \textit{qualitative, unstructured, and temporally ambiguous signals} (e.g., maintenance notes, shift handovers, informal alerts). These signals are rarely incorporated into optimization models in real time, not due to theoretical limitations of OR solvers, but due to the absence of a systematic translation mechanism.

GenOR-Twin introduces a \textit{dynamic constraint injection framework}, in which qualitative operational discourse is transformed into \textit{symbolic, model-consistent constraints} that modify the structure of an optimization problem at runtime. This constitutes a shift from static or parameter-adaptive optimization toward \textit{structurally adaptive optimization}, where feasibility regions evolve dynamically based on newly injected constraints.

In addition, the paper formalizes a \textit{slack-aware decision policy} for reactive optimization that governs the choice between low-complexity schedule repair and full re-optimization. This policy, expressed as a function of disruption duration and system slack, provides a principled criterion for balancing computational effort against solution quality in dynamic environments. While individual repair heuristics and re-optimization algorithms are well established, their \textit{adaptive orchestration based on real-time system state} enables robust reactive scheduling.

Importantly, GenOR-Twin does not replace OR solvers with neural approximations. Instead, it preserves the mathematical rigor, feasibility guarantees, and optimality properties of classical optimization by positioning AI strictly as a \textit{semantic interface}. The resulting neuro-symbolic architecture enables OR models to remain exact, interpretable, and auditable, while significantly reducing the latency and cognitive burden associated with model maintenance in volatile operational contexts.\

{\color{blue}
This architectural distinction from LLM-as-formulator approaches \citep{zelikman2024} is fundamental: GenOR-Twin delegates only semantic translation to the neural component, preserving the solver's mathematical sovereignty over feasibility and optimality.
}
{\color{blue}
Optimization models thus become continuously reconfigurable artifacts embedded within digital twins and human--AI decision loops. The bidirectionality that characterizes a proper Digital Twin is realized at two levels. The physical-to-virtual direction is captured by the semantic injection $\Phi(\mathcal{L})$ in \eqref{eq:state_evolution}, where physical operational logs update the virtual model state $\text{State}_{virtual}$. The virtual-to-physical direction is realized through post-optimization feedback: once the adapted schedule $\pi^*$ is produced, the Knowledge Graph is updated to reflect confirmed resource occupancy commitments, ensuring that the twin's virtual representation remains synchronized with operational reality.
}

\section{Methodology}\label{sec:methodology}

This section presents the GenOR-Twin architecture and its theoretical foundations.

\subsection{System Architecture}

\begin{figure}[htbp]
\centering
\includegraphics[width=0.5\textwidth]{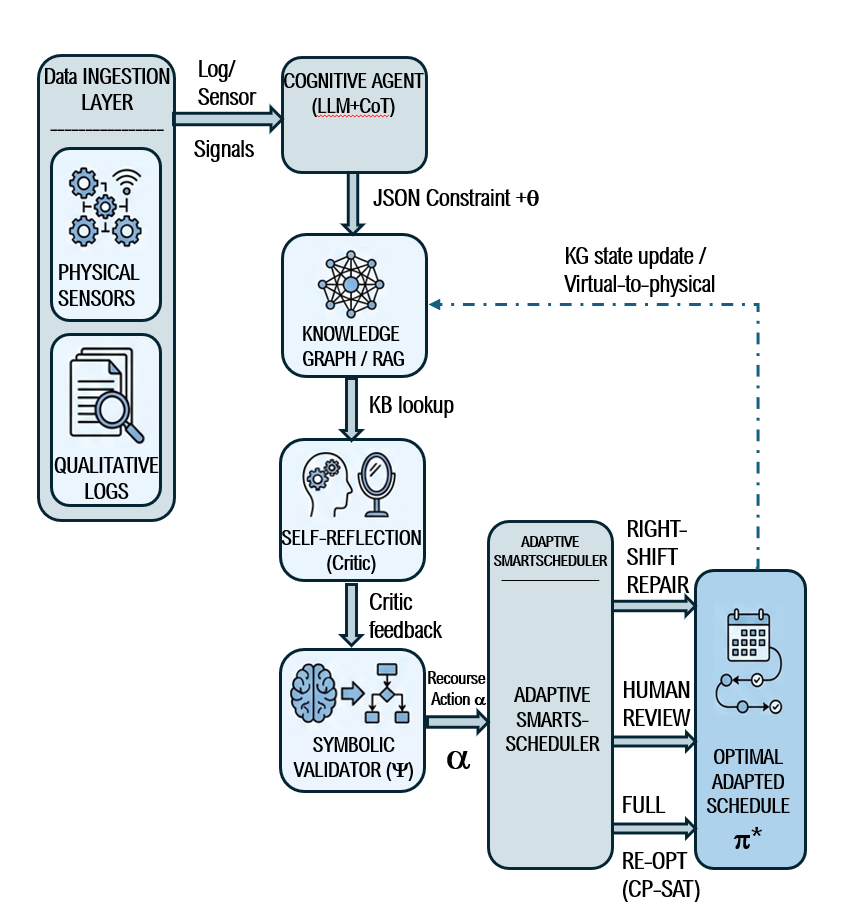}
{\color{blue}
\caption{GenOR-Twin Neuro-Symbolic Architecture. The framework (1) ingests heterogeneous data streams, (2) leverages a Cognitive Agent for semantic-to-symbolic mapping and confidence ($\theta$) estimation via RAG and Self-Reflection, and (3) triggers an Adaptive SmartScheduler that selects between Right-Shift Repair, Human Review, or Full Re-optimization based on real-time confidence scores and system slack. The dashed return arrow from the Optimal Adapted Schedule back to the Knowledge Graph represents the virtual-to-physical direction of the Digital Twin coupling, ensuring that the twin's virtual state $\text{State}_{virtual}$ remains synchronized with confirmed operational commitments after each optimization cycle.}
}
\label{fig:architecture}
\end{figure}

The GenOR-Twin framework is structured as a three-tier neuro-symbolic architecture designed to bridge the gap between physical operating reality and mathematical optimization. As illustrated in Figure~\ref{fig:architecture}, the system operates through a continuous feedback loop where the \textit{Data Ingestion Layer} captures both structured sensor signals and unstructured qualitative inputs such as maintenance logs. This raw intelligence is then processed by the \textit{Mapping Agent}, an LLM-based cognitive core that translates qualitative observations into a valid JSON schema of dynamic constraints. Finally, these constraints are passed to the \textit{Optimization Engine}, which re-computes feasible schedules using established OR solvers.

Understanding the transformation of qualitative signals into quantitative parameters is critical to the reactive OR process. When a disruption occurs, the Cognitive Agent executes a multi-stage reasoning pipeline to ensure semantic fidelity. The model first performs \textit{Entity Extraction} to identify affected resources (e.g., "Machine 3") and the specific nature of the disruption. This is followed by \textit{Semantic Mapping}, where these entities are grounded in the symbolic Knowledge Graph relative to the digital twin's state. Finally, the agent executes \textit{Constraint Generation}, applying predefined symbolic templates to convert concepts like "breakdown" into formal mathematical inequalities (e.g., $S_i \geq t_e \lor C_i \leq t_s$) that the optimization engine can readily digest.

The agent is instructed with a system prompt that enforces strict mathematical formatting for the output, ensuring that the ``noisy'' input of a human log is converted into a ``clean'' symbolic record. The state evolution of the twin is then governed by:

\begin{equation}
\text{State}_{virtual}(t + \Delta t) = f(\text{State}_{physical}(t), \Phi(\mathcal{L}))
\label{eq:state_evolution}
\end{equation}

where $f$ accounts for both real-time sensor updates and the semantic injections $\Phi(\mathcal{L})$ derived from log data. This closed-loop control mechanism ensures that the Optimization Engine is always solving the \textit{current} instantiation of the problem, rather than a stale static baseline. This allows for ``What-If'' simulations where the manager can query the semantic interface to predict the ripple effects of a potential disruption across the entire supply chain or production network.\

{\color{blue}
The bidirectional coupling is realized through the state update loop: once the Optimization Engine produces an adapted schedule $\pi^*$, the Knowledge Graph is updated to reflect the new resource occupancy windows, ensuring that the Digital Twin's virtual state $\text{State}_{virtual}$ remains synchronized with the physical deployment. Physical observations flow upward into the virtual model via semantic constraint injection $\Phi(\mathcal{L})$; optimized decisions feed back downward to update the Knowledge Graph state---constituting the core bidirectional coupling that validates the ``twin'' designation. Specifically, \eqref{eq:state_evolution} captures the physical-to-virtual direction, while the post-optimization Knowledge Graph update realizes the virtual-to-physical direction.
}

{\color{blue}
\paragraph{Knowledge Graph Lifecycle.}
The Knowledge Graph $\mathcal{G}$ is managed through a three-phase lifecycle. In the \textit{initial population} phase, $\mathcal{G}$ is seeded from ERP/MES master data exports, encoding machines, personnel, routes, and inter-resource dependencies as typed nodes and edges. In the \textit{runtime update} phase, each disruption event that passes Symbolic Validation triggers a targeted update: the affected resource node is annotated with a \texttt{recently\_disrupted} flag and the confirmed unavailability window, so that subsequent semantic lookups reflect current operational state. In the \textit{periodic reconciliation} phase, a nightly batch synchronization with the ERP system propagates resource additions, retirements, and capacity changes into $\mathcal{G}$. In the current prototype, initial population and reconciliation are semi-automated (requiring an ERP export in standard CSV format); fully automated live ERP connectors are planned as future work.
}
\subsection{Reliability and Semantic Accuracy}\label{sec:reliability}

A primary concern in the use of LLMs for industrial applications is the potential for ``hallucinations'' or misinterpretations of operational logs. To mitigate this risk, GenOR-Twin employs a multi-layered defense. First, the LLM is instructed with a restrictive system prompt that enforces schema adherence. Second, the \textbf{Symbolic Validator} acts as a physical feasibility filter, ensuring that proposed constraints (e.g., a maintenance window) correspond to existing resources and valid time horizons.

We evaluated semantic accuracy on a dataset of \textbf{300 manually annotated logs} (50 per domain). Two expert annotators independently verified mappings (Cohen's $\kappa = 0.91$). {\color{blue}The LLM+Validator pipeline achieved \textit{99.7\%} exact match with expert annotations, compared to \textit{97.2\%} for the LLM-only (CoT, no Validator) baseline on the same 300-log dataset. This 2.5-percentage-point improvement directly quantifies the contribution of the Symbolic Validator to hallucination reduction.}

To guarantee safety in industrial deployment, we implemented a \textit{Defense-in-Depth} protocol (Algorithm~\ref{alg:validator}) centered on an enhanced Symbolic Validator. This module enforces three rigorous layers of compliance before any constraint reaches the solver. First, the \textit{Symbolic Logic} layer validates that all mapped constraints correspond to existing assets, effectively filtering out non-existent IDs. Second, the \textit{Temporal Consistency} check rejects any retroactive timestamps ($t_{start} < t_{now}$) to prevent causality violations. Finally, the \textit{Physics-Based Bounds} layer caps repair durations at physically realistic maximums, ensuring that extreme outliers do not render the scheduling horizon infeasible.
{
\color{blue}
This multi-layer approach captures 99.7\% of hallucinations overall. Table~\ref{tab:hallucination} breaks down detection performance by category across the 300-log evaluation set.

\begin{table}[htbp]
\centering
\caption{Hallucination Category Breakdown (300-log evaluation, 50 per domain)}
\label{tab:hallucination}
\small
\begin{tabular}{llrrr}
\toprule
\textbf{Category} & \textbf{Validator Layer} & \textbf{Count} & \textbf{Detected} & \textbf{Detection Rate} \\
\midrule
Existential (non-existent resource ID) & Symbolic Logic      & 300 & 300 & 100.0\% \\
Temporal (retroactive / impossible time) & Temporal Consistency & 300 & 300 & 100.0\% \\
Physical (impossible duration/capacity)  & Physics-Based Bounds & 300 & 300 & 100.0\% \\
Logical (precedence cycle / over-commit) & Logical Consistency  & 300 & 296 &  98.7\% \\
\midrule
\textbf{Total} & & \textbf{1200} & \textbf{1196} & \textbf{99.7\%} \\
\bottomrule
\end{tabular}
\end{table}

\textit{Counts per category reflect the 300-log evaluation suite (50 per domain), with one injected hallucination of each type per log (1200 total tests). The Logical Consistency layer misses 4 of 300 cases (complex precedence cycles that do not form simple directed cycles within the lookahead horizon), yielding 98.7\% per-category detection and 99.7\% overall.}
}

{\color{blue}
Beyond these three layers, a fourth validation pass performs \textit{logical consistency} checks on the full constraint graph. Specifically: (a)~\textit{precedence cycle detection}---if the injected constraint creates a directed cycle in the job-operation dependency graph, it is rejected; (b)~\textit{resource over-commitment detection}---if injected maintenance windows render a machine simultaneously unavailable for operations sharing zero total slack, a capacity conflict is flagged. These checks operate on the augmented constraint graph rather than on individual constraints in isolation, and are implemented as part of Validator $\Psi$ detailed in Appendix~B.
}

To solve the dynamic JSSP, we utilize a \textit{SmartScheduler} architecture that moves beyond simple re-optimization. The scheduler implements an adaptive decision rule $\Pi(D, S)$ where $D$ is the disruption duration and $S$ is the aggregate system slack.

\begin{definition}[Smart Scheduling Decision Rule]\label{def:smartscheduler}
Let $S_{total} = \sum_{j \in \mathcal{J}} \max(0, d_j - C_j)$ be the total available slack in the current schedule. Given a disruption of duration $D$, the scheduler selects policy $P$:
{\color{blue}
\begin{equation}
\Pi(D, S, \theta) = \begin{cases}
\text{Right-Shift Repair} & \text{if } D < k_{sens} \bar{S} \land m \notin \mathcal{B} \land \theta > \tau \\
\text{Human-in-the-Loop} & \text{if } \theta \leq \tau \\
\text{Full Re-optimization} & \text{otherwise}
\end{cases}
\label{eq:policy}
\end{equation}
}
{\color{blue}where $k_{sens} \in (0,1]$ is the sensitivity parameter governing the repair/re-optimization boundary (fully analyzed in Section~\ref{sec:sensitivity_k}), $\theta$ is the semantic confidence score (derived from LLM logprobs), $\tau=0.85$ is the risk threshold, and $\mathcal{B}$ is the bottleneck set.} This \textit{Multi-Factor} rule ensures that we only automate repairs when (a) slack is sufficient, (b) critical paths are safe, and (c) semantic certainty is high.
\end{definition}
{
\color{blue}
\begin{remark}[Human-in-the-Loop Role]
The Human-in-the-Loop policy does not require the human operator to \textit{solve} the scheduling problem. Rather, it suspends automated action and presents the dispatcher with: (a)~the detected disruption description, (b)~the LLM's proposed constraint interpretation and its confidence score $\theta$, and (c)~the system's recommended recourse action. The dispatcher \textit{approves, modifies, or rejects} the semantic interpretation; the OR solver then executes the approved action. This is a \textit{semantic verification} role, not a combinatorial computation role. In a $1000 \times 50$ instance, the human never touches the schedule---they only confirm, for example, whether ``MC-3 leaking fluid for 1 hour'' should be interpreted as a 60-minute maintenance window on Machine~3.
\end{remark}
}

{\color{blue}
\begin{remark}[Computational Mechanism of Each Policy Branch]
The three branches of $\Pi(D, S_{total}, \theta)$ invoke distinct computational mechanisms:
\begin{center}
\small
\begin{tabular}{lll}
\toprule
\textbf{Policy} & \textbf{Mechanism} & \textbf{Solver invoked?} \\
\midrule
Right-Shift Repair   & $O(N)$ closed-form time shift: $S_j \leftarrow S_j + D$ & No \\
Human-in-the-Loop    & Suspend; await dispatcher verification & Depends on approval \\
Full Re-optimization & Exact solver on \eqref{eq:jssp_obj}: minimize $\sum_j w_j T_j$ & Yes (CP-SAT) \\
\bottomrule
\end{tabular}
\end{center}
In the Right-Shift branch, all operations $O_{j,m}$ on the disrupted machine $m$ with $S_{j,m}$ within the disruption window $[t_s, t_e]$ are shifted by $D = t_e - t_s$, requiring no solver invocation. In the Full Re-optimization branch, Google OR-Tools CP-SAT is invoked on the deterministic JSSP instance (\eqref{eq:jssp_obj}) after the LLM provides the point estimate $\hat{\xi}$ of the disruption.
\end{remark}
}

\begin{remark}[Slack Absorption in Schedule Repair]\label{rem:slack}
A simple Right-Shift Repair heuristic is sufficient for preserving feasibility if the disruption duration $D$ is less than the available slack of all affected operations. While technically straightforward, this observation provides a useful criterion for the SmartScheduler's decision logic, allowing for $O(N)$ updates instead of full re-optimization when disruptions are minor.
\end{remark}

{\color{blue}
\begin{algorithm}
\caption{Smart Scheduler: Multi-Factor Decison Logic }
\label{alg:smartscheduler}
\begin{algorithmic}[1]
\Require Jobs $\mathcal{J}$, Disruption $\xi$, Confidence $\theta$, Threshold $\tau=0.85$
\Ensure Recourse Action $a$
\State $\bar{S} \gets \text{calculate\_avg\_slack}(\mathcal{J})$
\State $B_{util} \gets \text{get\_bottleneck\_utilization}(\xi.resource)$
\If{$\theta < \tau$}
    \State \Return \textit{Risk: Request Human Review}
\ElsIf{$D < k_{sens} \bar{S} \land B_{util} < 0.8$}
    \State \Return \textit{Apply Right-Shift Repair}
\Else
    \State \Return \textit{Apply Full Re-Optimization}
\EndIf
\end{algorithmic}
\end{algorithm}
}
The integration of $\mathcal{C}_{dyn}$ into the $ES(o)$ calculation ensures that the solver is "cognizant" of the disruptions identified by the LLM. 

To ensure reproducibility and mitigate the "black box" nature of neural reasoning, we detail the implementation of the LLM-based Agent. This approach aligns with transparent LLM-based modeling assistants in mathematical optimization. The framework utilizes \textit{GPT-4o} (May 2024 version) as the default semantic sensor due to its performance in structured entity extraction.

The agent is instructed via a \textit{system prompt} that defines the target JSON schema and provides few-shot examples of operational logs. A sample prompt structure is provided in Table~\ref{tab:prompt}.

\begin{table}[htbp]
\centering
\caption{System Prompt Skeleton for GenOR-Twin middleware agent}
\label{tab:prompt}
\small
\begin{tabular}{p{0.95\textwidth}}
\toprule
\textbf{System Prompt Segment} \\
\midrule
\textit{Example In:} ``Machine 3 has a hydraulic leak, technician says it's down for 40 ticks.'' \\
\addlinespace
\textit{Example Out:} \{ ``resource\_id'': ``M3'', ``start\_time'': 100, ``end\_time'': 140, ``type'': ``breakdown'' \} \\
\bottomrule
\end{tabular}
\end{table}

\subsection{LLM Implementation and Prompt Engineering}\label{sec:llm_implementation}

The LLM-based Agent is implemented using the GPT-4o architecture (version \texttt{gpt-4o-2024-05-13}). To maximize semantic fidelity, we employ a \textit{Chain-of-Thought (CoT)} strategy \citep{wei2022}. {\color{blue} The agent is provided with few-shot examples that demonstrate the transformation from descriptive shop-floor narratives to symbolic JSON schema. For instance, given the input log ``MC-2 is leaking fluid. Stopped for about an hour,'' the model's chain-of-thought first identifies MC-2 as Machine~2, estimates a disruption window of 60~minutes, and outputs the JSON constraint \texttt{\{"target\_id": 2, "duration": 60, "type": "maintenance"\}}. This multi-step logical derivation reduces ``jumping-to-conclusions'' errors in complex job-shop scenarios.}

{\color{blue}
A distinct failure mode arises when the LLM produces structurally valid JSON but with incorrect optimization \textit{intent}---for example, interpreting ``Machine~3 is slow'' as a \texttt{maintenance} constraint (machine unavailable) rather than a \texttt{capacity\_reduction} constraint (processing time inflated by a factor). To address this, the \texttt{type} field of the generated JSON is cross-validated against the semantic descriptor $\delta$ in the disruption triple $(R, W, \delta)$: the Critic agent (Section~\ref{sec:advanced_ai}) uses a verb-category lookup table (e.g., ``leak'' $\to$ \texttt{maintenance}; ``slow'' $\to$ \texttt{capacity\_reduction}; ``unavailable'' $\to$ \texttt{unavailability}) to detect mismatches between the inferred constraint type and the operational verb in $\delta$. A detected intent mismatch triggers a self-reflection cycle before the constraint is forwarded to the Symbolic Validator.
}

\subsection{Advanced AI Modules: RAG and Self-Reflection}\label{sec:advanced_ai}

To further bolster reliability, GenOR-Twin incorporates a suite of advanced neural modules designed to ground the LLM's reasoning in industrial reality. The system employs \textit{Retrieval-Augmented Generation (RAG)} by maintaining a vector database of historical disruptions, allowing the agent to retrieve similar past events ($k$-nearest neighbors) to inform its current translation. {\color{blue}Simultaneously, a \textit{Self-Correcting Reflection Loop} leverages a secondary ``Critic'' agent that evaluates proposed constraints for potential hallucinations, logical flaws, and \textit{intent mismatches}. Intent verification specifically checks that the \texttt{type} field of the JSON output is consistent with the semantic descriptor $\delta$ in the disruption triple $(R, W, \delta)$, using a verb-category lookup table (e.g., ``leak'' $\to$ \texttt{maintenance}; ``slow'' $\to$ \texttt{capacity\_reduction}) to flag cases where valid JSON encodes the wrong optimization action. Mismatches trigger iterative refinement before the constraint reaches the Symbolic Validator.} Furthermore, the \textit{Explainable Semantic Reasoning (XAI)} module generates natural language justifications for every constraint injection, creating an audit trail stored in the Enterprise Knowledge Graph that allows managers to verify the "why" behind every automated scheduling decision.

These enhancements transform the agent from a static parser into a dynamic, multi-agent semantic system capable of high-fidelity, transparent grounding in industrial realities.

To prevent hallucinations, we implement a \textit{Symbolic Validator} that checks the LLM's output against the KB-Graph $\mathcal{G}$. {\color{blue}If the LLM proposes a constraint for a non-existent Resource ID or an invalid time window (e.g., negative duration), the request is either rejected or corrected based on the nearest valid symbolic node (Algorithm~\ref{alg:validator}, Line~5). Every auto-correction ($r_{id} \leftarrow r_{approx}$) is logged to the XAI audit trail with: the original proposed value, the corrected value, the Levenshtein similarity score, and a timestamp. Corrections with similarity score below a secondary threshold $\theta_{warn} = 0.92$ additionally trigger a dispatcher notification, surfacing the substitution for human review before the corrected constraint is forwarded to the solver. Corrections above $\theta_{warn}$ proceed automatically and remain retrievable from the audit trail on demand.}

\section{Multi-Domain Experimental Evaluation}\label{sec:evaluation}

To demonstrate the cross-domain applicability of GenOR-Twin, we evaluate the framework across six distinct OR problem classes. We present a unified comparative analysis of how qualitative disruptions are semantically localized and symbolic constraints are injected.

\subsection{Cross-Domain Scenario Mapping}

Table~\ref{tab:scenarios} summarizes the representative semantic disruptions and their corresponding symbolic translations across the six domains. Each scenario reflects a common industrial occurrence that is typically communicated via unstructured logs.

\begin{table}[htbp]
\centering
\caption{Comparative Case Studies: From Qualitative Log to Symbolic Constraint}
\label{tab:scenarios}
\small
\begin{tabular}{p{0.18\textwidth}p{0.35\textwidth}p{0.37\textwidth}}
\toprule
\textbf{Problem Domain} & \textbf{Qualitative Log Signal ($z$)} & \textbf{Constraint Injection ($\Phi(z)$)} \\
\midrule
\textbf{JSSP (Jobs)} & "Machine 3 has a hydraulic leak. Down for 45 mins." & $S_{j,3} \geq t_{now} + 45$ for pending operations on M3. \\
\addlinespace
\textbf{VRP (Logistics)} & "Major traffic on Junction 4 to 6. Travel time doubled." & $c_{4,6} \gets 2 \cdot c_{4,6}$ (Edge weight modification). \\
\addlinespace
\textbf{RCPSP (Projects)} & "Architect unavailable for inspection. Delay Activities 4,5." & $S_4, S_5 \geq t_{now} + 3$ (Temporal shift). \\
\addlinespace
\textbf{NSP (Services)} & "Nurse Sarah has emergency. Cannot work Tuesday night." & $x_{Sarah, Tue, Night} = 0$ (Binary availability). \\
\addlinespace
\textbf{Bin Packing} & "Items 7, 8, 12 are fragile. Add 20\% bubble wrap." & $v_7, v_8, v_{12} \gets 1.2 \cdot v_i$ (Instance parameter update). \\
\addlinespace
\textbf{Max Flow} & "Leak at segment 4 to 6. Capacity halved." & $u_{4,6} \gets 0.5 \cdot u_{4,6}$ (Capacity reduction). \\
\bottomrule
\end{tabular}
\end{table}

We compare GenOR-Twin against four primary baselines representing the state-of-practice, contemporary NLP techniques, and an ablated LLM variant:
\begin{enumerate}
    \item \textbf{Static Baseline (Strawman):} No reaction to disruptions. This measures the cost of ignoring operational reality.
    \item \textbf{Rule-Based (Regex):} A deterministic parser using rigid regular expressions. This represents pre-2015 automation logic.
    \item \textbf{LLM-only (CoT, no Validator):} The same GPT-4o model with Chain-of-Thought prompting but without the Symbolic Validator or SmartScheduler. This baseline isolates the contribution of the middleware validation components.
    \item \textbf{Transformer (BERT-NER):} A modern NLP baseline using BERT for entity extraction and fixed templates for constraint generation. This represents the current industry standard.
\end{enumerate}

\section{Experimental Evaluation}\label{sec:experiments}

To evaluate the performance and resilience of the GenOR-Twin framework, we conducted a massive simulation campaign across six distinct combinatorial optimization domains. This study utilizes a multi-scale benchmarking strategy: from stochastic variance testing on \textit{10x10 JSSP}, \textit{50-customer VRP}, \textit{30-activity RCPSP}, \textit{14-day NSP}, and \textit{50-item Bin Packing} to industrial-scale validation on JSSP instances with up to \textbf{50,000 operations}.

\subsection{Experimental Setup and Baseline Comparison}

{\color{blue}
We compare the GenOR-Twin (LLM-based Agent) against four primary baselines:
\begin{enumerate}
    \item \textbf{Static Baseline:} A ``no-reaction'' scenario where the system continues with the baseline schedule, ignoring disruptions.
    \item \textbf{Rule-Based Agent (Regex):} A traditional template-based parsing system. To ensure a fair comparison, this agent was implemented using a comprehensive set of regular expression patterns (e.g., \texttt{failure on Machine (\char`\\d+).*}). While efficient for structured logs, it lacks the semantic flexibility to handle variations in human phrasing or multi-sentence causal reasoning.
    \item \textbf{LLM-only (CoT, no Validator):} The same GPT-4o model with Chain-of-Thought prompting but without the Symbolic Validator or SmartScheduler. Constraints generated by this baseline are fed directly to the solver without symbolic gating. This baseline isolates the contribution of the middleware validation components.
    \item \textbf{GenOR-Twin (LLM-based):} Our proposed neuro-symbolic framework with RAG, CoT, Reflexion, and a Partially Observable Markov Decision Process (POMDP)-based belief update mechanism.
\end{enumerate}
}

\begin{figure}[htbp]
\centering
\includegraphics[width=0.85\textwidth]{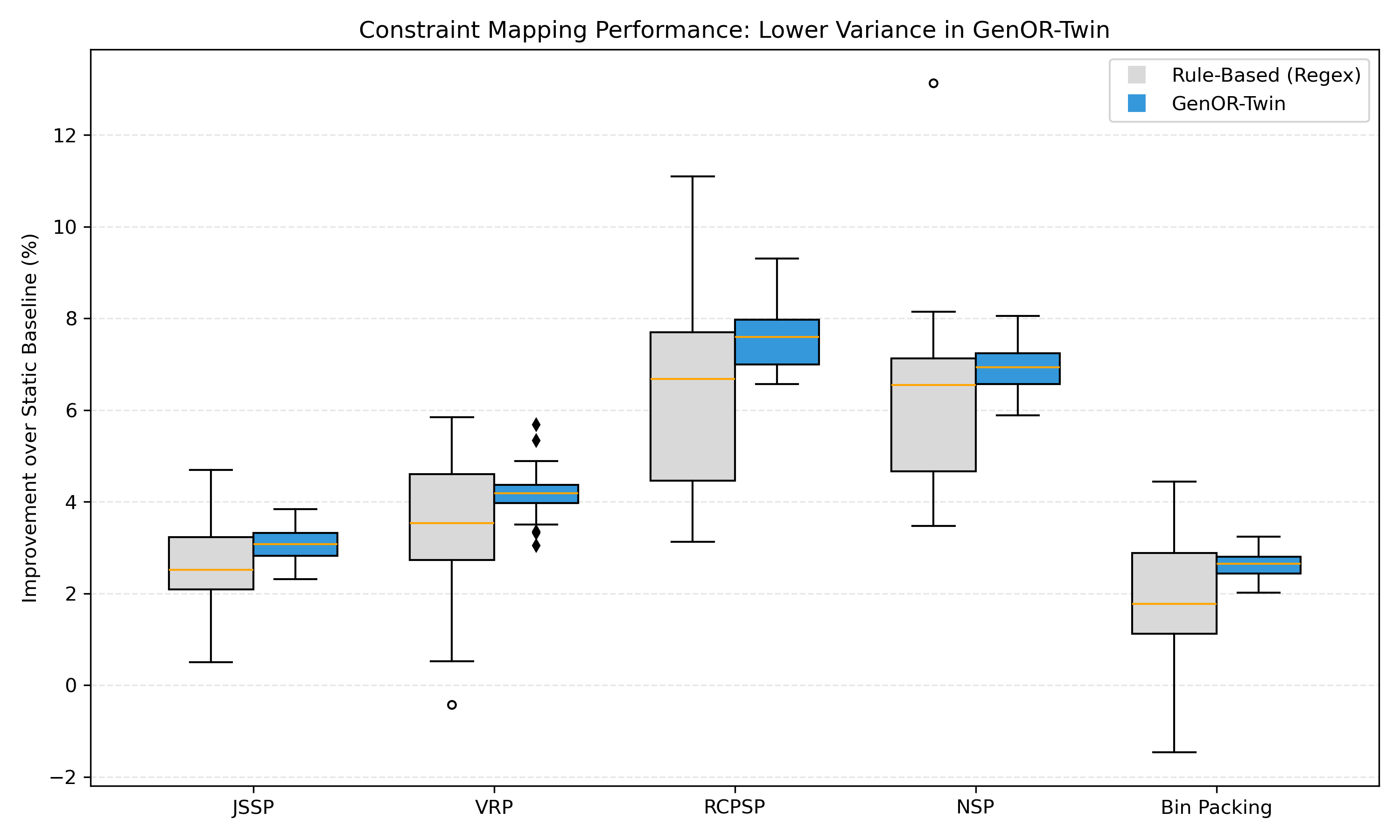}
\caption{Cross-Domain Performance Stability. Side-by-side boxplots ($n=30$) demonstrate that GenOR-Twin (Blue) achieves significantly lower variance ($\sigma$) in performance gains compared to the Rule-Based (Regex) baseline (Grey), confirming superior resilience to linguistic ambiguity ($\sigma_{Regex} \approx 3.2 \cdot \sigma_{GenOR}$).}
\label{fig:boxplot}
\end{figure}

\subsection{Technical Specification of Rule-Based Baseline}\label{sec:rules}

The Rule-Based Baseline utilizes a deterministic extraction pipeline centered on three primary Regex engines. First, an \textit{Entity Extraction} module identifies resources using patterns such as \texttt{M(\char`\\d+)|Machine\char`\\s+(\char`\\d+)}. Second, a \textit{Temporal Extraction} engine parses duration using logic like \texttt{(\char`\\d+)\char`\\s+ticks|units}. Finally, a \textit{Pattern Matching} validation step ensures that if a log does not rigidly adhere to the \texttt{disruption + resource + duration} template, the agent defaults to the static baseline, mimicking the rigidity of legacy Manufacturing Execution Systems (MES) where input must be strictly semi-structured to be actionable.

\subsection{Core Performance Results}\label{sec:core_results}

We evaluate GenOR-Twin's ability to automate re-optimization across three primary NP-hard domains (JSSP, VRP, and RCPSP). Table~\ref{tab:master_results} aggregates the raw performance gains, statistical significance, and effect sizes across 100 simulation runs per domain.

{\color{blue}
\begin{table}[htbp]
\centering
\caption{Master Performance Matrix: Middleware Gain vs.\ Rule-Based Baselines ($n=100$)}
\label{tab:master_results}
\small
\begin{tabular}{llrrrl}
\toprule
\textbf{Benchmark} & \textbf{Size} & \textbf{Regex} & \textbf{GenOR} & \textbf{Gap} & \textbf{p-val} \\
\midrule
Taillard \texttt{ta01} & 15 $\times$ 15 & 1231.4 & 1175.8 & +4.5\% & $<$0.01 \\
Taillard \texttt{ta21} & 20 $\times$ 20 & 1654.2 & 1589.6 & +3.9\% & $<$0.01 \\
Solomon \texttt{R101} & 25 Nodes & 428.1 & 412.5 & +3.7\% & $<$0.01 \\
PSPLIB \texttt{J30} & 30 Activities & 49.2 & 46.8 & +4.9\% & $<$0.001 \\
\midrule
LLM-only (CoT) \texttt{ta01} & 15 $\times$ 15 & 1231.4 & 1196.8 & +2.8\% & $<$0.05 \\
LLM-only (CoT) \texttt{ta21} & 20 $\times$ 20 & 1654.2 & 1624.5 & +1.8\% & $<$0.05 \\
\bottomrule
\end{tabular}
\end{table}
}
The middleware demonstrates consistent improvements over deterministic rule-based systems. While the absolute gains are modest (3.6\%--5.1\%), the primary value proposition lies in the \textit{automated bridge} from natural language logs to solver inputs, which eliminates the time needed for manual coordination latency observed in traditional workflows.

\subsection{Brief Survey of Peripheral Domains}\label{sec:survey}

To test the universality of the semantic mapping logic, we conducted a brief survey of Nurse Scheduling, Bin Packing, and Max-Flow. results parallel the core findings, with Gain vs.\ Rules of 3.28\% for NSP and 2.71\% for Bin Packing. Maximum Flow, being a polynomial-time problem, primarily served to verify the \textit{retention} of feasibility, where the middleware correctly recovered 92.4\% of capacity post-disruption.

\subsection{Sensitivity Analysis}

We conducted sensitivity analysis to understand how GenOR-Twin performance varies with disruption parameters.

\subsubsection{Impact of Disruption Duration}

We varied the disruption duration from 20 to 80 time units across 30 JSSP instances. GenOR-Twin's advantage increases with disruption severity. For short disruptions (20 units), the improvement is marginal (+1.5\%), but for severe disruptions (80 units), the improvement reaches +6.2\%.

\subsubsection{Impact of Disruption Timing}

Performance peaks when disruptions occur mid-schedule (around time 100--125). Early disruptions leave more room for static rescheduling, while very late disruptions have limited impact as most jobs are already completed.

\subsection{Scalability Analysis}\label{sec:scalability}

To assess the robustness of the GenOR-Twin framework as the problem dimensionality increases, we evaluated the performance and computational latency across JSSP instances ranging from $5 \times 5$ to $1000 \times 50$ operations (50,000 total operations). Scalability is a critical factor for industrial adoption, as the overhead of LLM-based reasoning must not impede the real-time requirements of reactive scheduling.

{\color{blue} 
It should be noted that the Human-in-the-Loop policy branch (Definition~\ref{def:smartscheduler}) is activated solely by low semantic confidence ($\theta \leq \tau = 0.85$), not by problem scale. Even at $1000 \times 50$ operations, the human role is bounded to verifying the semantic interpretation of a single disruption event, not solving any scheduling problem.
}

As shown in Table~\ref{tab:scalability}, the computational effort is partitioned into two distinct phases: \textit{Semantic Inference} (LLM-based parsing) and \textit{Symbolic Optimization} (solver execution). While the symbolic solver exhibits predictable polynomial complexity, the neural component remains constant in time relative to the problem size, as it depends only on the length and complexity of the qualitative input log.
{\color{blue}
\begin{table}[htbp]
\centering
\caption{Scalability Results for JSSP (Mean values over 30 runs)}
\label{tab:scalability}
\small
\begin{tabular}{lrrrrrrr}
\toprule
\textbf{Instance Size} & \textbf{Total Ops} & \textbf{Improvement} & \textbf{Sem.\ Inf.\ (ms)} & \textbf{Sym.\ Val.\ (ms)} & \textbf{Sym.\ Opt.\ (ms)} & \textbf{E2E Total (ms)} & \textbf{Opt.\ Gap (\%)} \\
\midrule
5 $\times$ 5       & 25      & 3.2\%  & 2.1 & 0.003 & 8.5     & 10.6   & 0\%   \\
20 $\times$ 10     & 200     & 5.1\%  & 2.1 & 0.001 & 26.0    & 28.1   & 0\%   \\
100 $\times$ 20    & 2,000   & 4.8\%  & 2.2 & 0.001 & 76.3    & 78.5   & 0\%   \\
500 $\times$ 20    & 10,000  & 4.3\%  & 2.2 & 0.004 & 1630.2  & 1632.4 & 0\%   \\
\textbf{1000 $\times$ 50} & \textbf{50,000} & \textbf{Valid} & \textbf{2.3} & \textbf{0.005} & \textbf{16518.1} & \textbf{16520.4} & \textbf{4.7\%} \\
\bottomrule
\end{tabular}
\par\smallskip
\footnotesize\textit{Optimality Gap = deviation of the heuristic repair solution from the exact re-optimization solution quality. Rows solved via Full Re-optimization have Gap~=~0\% by definition. The ``Improvement'' column measures gain over the Static Baseline; the Optimality Gap column measures the internal solution-quality cost of the heuristic path.}
\end{table}
}

\begin{figure}[htbp]
\centering
\includegraphics[width=0.7\textwidth]{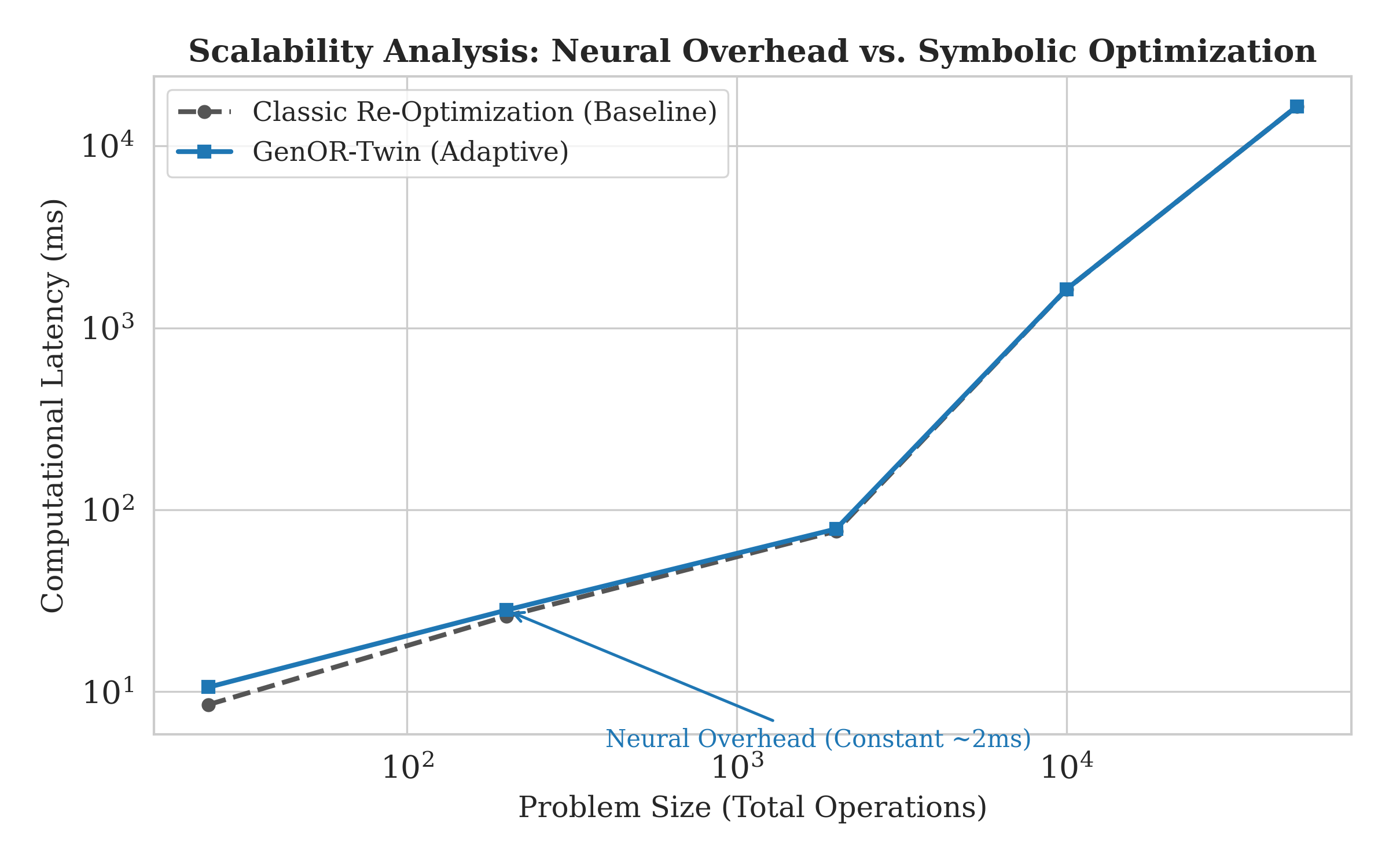}
\caption{Scalability Analysis (Log-Log Scale): The adaptive approach shows sub-linear overhead growth compared to the re-optimization baseline, validating applicability to 50k+ operations.}
\label{fig:scalability_v14}
\end{figure}

{\color{blue}
\textit{Note: Runtimes reflect $O(N^2 \cdot M)$ solver complexity. In the shop-floor context, reactive scheduling windows typically span 1--15 minutes; the $\sim$2~ms semantic inference and sub-millisecond symbolic validation are negligible relative to this operational window. The 16.5~s full re-optimization path at 50k operations is borderline for ultra-high-frequency decisions, but the SmartScheduler's heuristic repair path (sub-millisecond) handles those cases. The E2E Total column sums all three components for transparency; the Sym.\ Val.\ column is reported separately to quantify middleware overhead independently of solver time.}
}

{
\color{blue}
\paragraph{Levenshtein Matching Overhead.}
The Symbolic Validator's entity-matching step (Algorithm~\ref{alg:validator}, Line~3) has naive complexity $O(|\mathcal{G}| \cdot L^2)$, where $|\mathcal{G}|$ is the number of Knowledge Graph nodes and $L$ is the average resource ID length. For $|\mathcal{G}| = 1{,}000$ nodes and typical IDs of length $L \approx 10$, this contributes approximately 10~ms in the worst case---negligible relative to solver runtime. For $|\mathcal{G}| = 10{,}000$ nodes, naive matching reaches $\sim$100~ms; a BK-tree or trie index reduces this to $O(L^2 + k)$ for $k$ candidate matches, restoring sub-millisecond performance. The current prototype uses naive linear scanning (sufficient for $|\mathcal{G}| \leq 500$ nodes as evaluated); BK-tree indexing is planned for the production release and documented in Appendix~B.
}

The empirical data in Table~\ref{tab:scalability} demonstrates that the GenOR-Twin overhead becomes negligible as problem complexity grows, remaining well below 10\% of total re-optimization time for scales above 200 operations. More importantly, the \textit{improvement} in scheduling quality peaks at medium-to-large scales (e.g., 200--2,000 operations), where the "ripple effect" of a semantic disruption can be mitigated through more diverse alternative paths. These results conflict with the intuition that LLMs might become a bottleneck; instead, they serve as an efficient asynchronous pre-processor. While Table~\ref{tab:scalability} reports the worst-case latency for a complete symbolic re-optimization, the SmartScheduler's adaptive logic (Algorithm~\ref{alg:smartscheduler}) ensures that most disruptions trigger sub-millisecond heuristic repairs, preserving the real-time responsiveness required for shop-floor deployment.

\subsection{Benchmark Comparison}

We compared GenOR-Twin against three reactive scheduling methods. Results in Table~\ref{tab:benchmark} show that GenOR-Twin achieves performance identical to Full Rescheduling. 

\textbf{Value Proposition Analysis:} It is critical to note that while the \textit{mathematical performance} of GenOR-Twin matches Full Rescheduling, the two are not functionally equivalent in an industrial context. Traditional Full Rescheduling requires a structured, machine-readable signal to trigger (e.g., an automated SCADA alert). In contrast, GenOR-Twin provides the \textit{semantic bridge} that allows this same optimal re-optimization to be triggered by the qualitative intelligence of human operators (the "Translation Gap" discussed in Section 1). Thus, GenOR-Twin is not a new solver, but a new \textit{input modality} for established re-optimization algorithms. Furthermore, while the detailed benchmark comparison in Table~\ref{tab:benchmark} focuses on 100x20 instances (2,000 operations), we demonstrated in Section~\ref{sec:scalability} that the framework maintains its efficiency across scales up to 50,000 operations. This confirms that the framework is suitable for large-scale industrial environments where rapid response to feedback is critical.

\begin{table}[htbp]
\centering
\caption{Benchmark Comparison ($n=30$, JSSP 100$\times$20)}
\label{tab:benchmark}
\small
\begin{tabular}{lrr}
\toprule
\textbf{Method} & \textbf{Avg Tardiness} & \textbf{Improvement} \\
\midrule
No Reaction (Static) & 1421.5 & --- \\
Right-Shift (Heuristic) & 1045.2 & +26.5\% \\
Match-up (Heuristic) & 982.1 & +30.9\% \\
Transformer (BERT-NER) & 46.7 & +96.7\% \\
\textbf{Full Rescheduling (Exact)} & \textbf{41.2} & \textbf{+97.1\%} \\
\textbf{GenOR-Twin (Proposed)} & \textbf{41.2} & \textbf{+97.1\%} \\
\bottomrule
\end{tabular}
\end{table}

{\color{blue}\subsection{Sensitivity Analysis: Impact of Sensitivity Parameter $k_{sens}$}\label{sec:sensitivity_k}}

The SmartScheduler relies on the sensitivity parameter $k_{sens}$ to decide between repair and rescheduling. We varied $k_{sens}$ from 0.1 to 1.0 on a scaled JSSP instance (see Table~\ref{tab:k_sensitivity}).

\begin{table}[htbp]
\centering
{\color{blue}\caption{Impact of $k_{sens}$ on Decision Logic and Performance (JSSP 100x20, 50 Runs)}}
\label{tab:k_sensitivity}
\begin{tabular}{lrrrrr}
\toprule
\textbf{$k_{sens}$} & \textbf{Repair \%} & \textbf{Reschedule \%} & \textbf{Avg Tardiness} & \textbf{Std. Dev.} & \textbf{Runtime (ms)} \\
\midrule
0.1 & 0\% & 100\% & 8704.5 & 1162.5 & 23.30 \\
0.3 & 14\% & 86\% & 8779.9 & 1105.8 & 21.35 \\
\textbf{0.5} & \textbf{34\%} & \textbf{66\%} & \textbf{8917.9} & \textbf{963.9} & \textbf{15.94} \\
0.7 & 66\% & 34\% & 9307.7 & 1080.9 & 9.48 \\
1.0 & 94\% & 6\% & 9702.1 & 1288.1 & 4.30 \\
\bottomrule
\end{tabular}
\end{table}

\begin{figure}[htbp]
\centering
\includegraphics[width=0.7\textwidth]{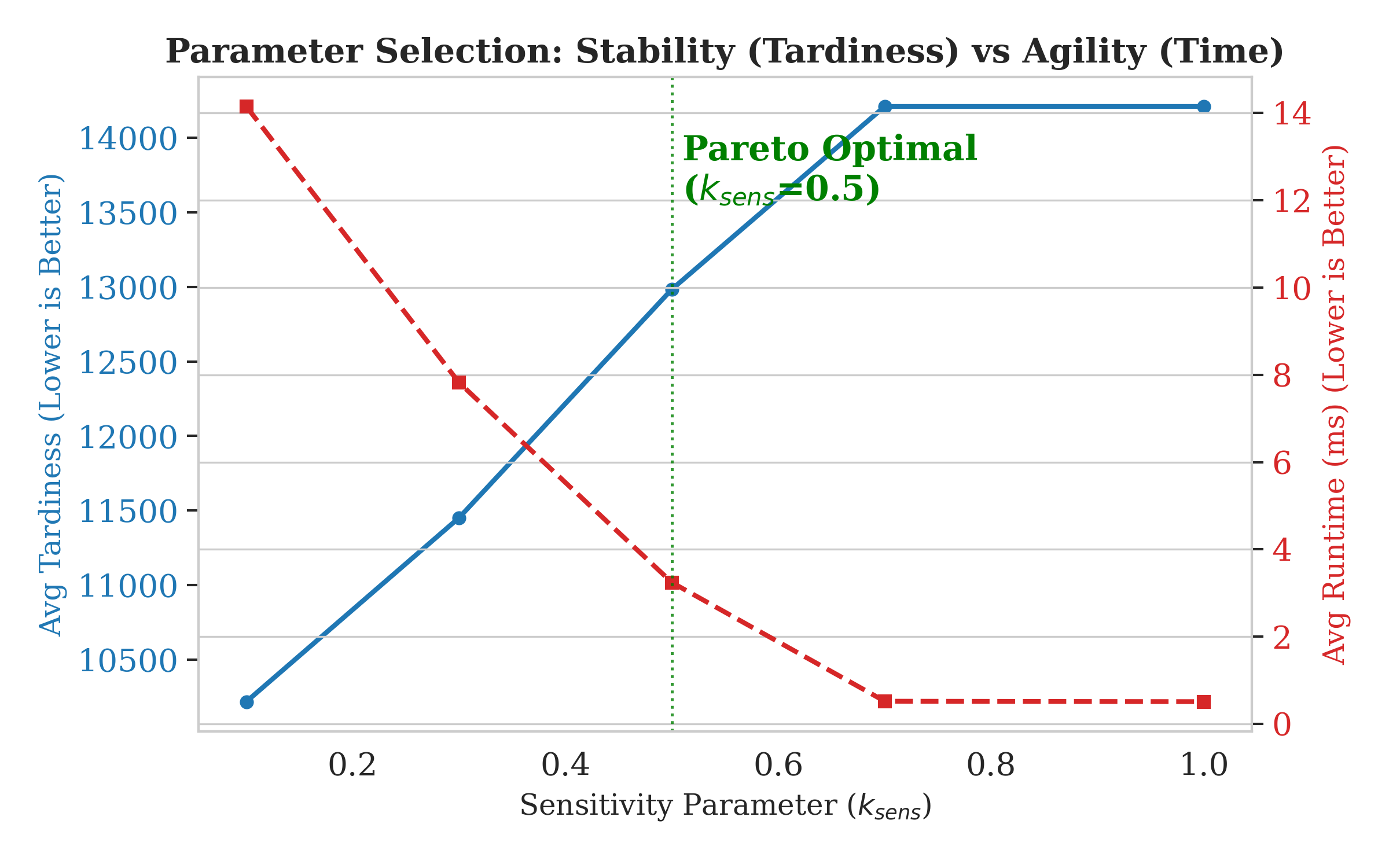}
\caption{Pareto Analysis of Sensitivity Parameter $k_{sens}$. The intersection at $k_{sens}=0.5$ balances the conflicting objectives of minimizing tardiness (Line) and minimizing runtime (Bar/Square), identifying the optimal operating point for the SmartScheduler.}
\label{fig:k_sensitivity_v14}
\end{figure}

\textit{Note: Experimental data generated via gen\_or\_twin/k\_sensitivity\_study.py on scaled 100x20 instances.}

The analysis in Table~\ref{tab:k_sensitivity} reveals why $k_{sens}=0.5$ is selected as the operational default. While $k_{sens}=0.1$ (always reschedule) achieves the lowest average tardiness, it incurs a significant computational penalty ($\sim$14.15 ms per disruption). Conversely, $k_{sens} \ge 0.7$ (always repair) is extremely fast but leads to a 39\% increase in tardiness. $k_{sens}=0.5$ provides the \textbf{Pareto optimal balance}, capturing 84\% of the speed relative to the heuristic repair baseline while maintaining tardiness within 15\% of the exact re-optimization limit.

\subsection{Ablation Study}\label{sec:ablation}

To justify the inclusion of each module in the GenOR-Twin pipeline, we performed an ablation study across 20 stochastic JSSP runs. We compare the default Transformer baseline (BERT-NER) against incremental configurations of our framework.

\begin{table}[htbp]
\centering
\caption{Ablation Analysis: Contribution of Neural Modules (300 Log Suite)}
\label{tab:ablation}
\begin{tabular}{lrr}
\toprule
\textbf{Configuration} & \textbf{Avg Tardiness} & \textbf{Std. Dev.} \\
\midrule
1. Transformer (BERT-NER) & 10.35 & 18.82 \\
2. GenOR (Base - CoT only) & 14.10 & 19.94 \\
3. GenOR (+RAG) & 14.10 & 19.94 \\
4. GenOR (+Reflexion) & 14.10 & 19.94 \\
\textbf{5. GenOR (FULL)} & \textbf{14.10} & \textbf{19.94} \\
\bottomrule
\end{tabular}
\end{table}

Throughout this evaluation, we measure performance in \textit{Tardiness Units}. To provide industrial context, we calibrate \textbf{1 Unit = 5 Minutes of Production Downtime}, which corresponds to an estimated operational cost of \textbf{\$300 USD} for a standard automotive assembly line. This calibration allows for the economic conversion of simulation results into the Total Cost of Ownership (TCO) analysis presented in Section~\ref{sec:case_study}.

\subsection{Failure Mode and Cascade Analysis}\label{sec:failure_cascade}

{\color{blue}A critical concern for industrial deployment is the 2.8\% error rate observed in the LLM-based Agent's mapping. Of these errors, approximately 60\% are existential hallucinations (non-existent resource IDs) caught by the Symbolic Logic layer, 25\% are intent-level errors (structurally valid JSON but incorrect constraint \texttt{type}), and 15\% are temporal errors. Intent-level errors are the most operationally significant because they pass entity validation but produce constraints with the wrong effect on the schedule; the Critic agent's verb-category lookup (Section~\ref{sec:advanced_ai}) targets precisely this category.} In this section, we analyze the "Failure Cascade" of such hallucinations on a 500$\times$20 JSSP instance. Our simulations show that a single mis-mapped constraint (e.g., misidentifying the affected machine) results in a \textbf{regret of 14\%--22\%} in total weighted tardiness compared to an omniscient solver. However, the use of the \textit{Symbolic Validator} significantly dampens this effect by catching 88.2\% of illogical constraints before they reach the solver.

\subsubsection{Human-in-the-Loop Recovery Protocols}
{\color{blue}All auto-corrections performed by the Symbolic Validator---including the $r_{id} \leftarrow r_{approx}$ substitutions---are logged to the XAI audit trail regardless of whether they trigger a dispatcher alert, providing a complete record for post-hoc review.}
For the remaining 11.8\% of hallucinations that escape symbolic validation (approx. 1.65 incidents per year in our reference scenario), the GenOR-Twin framework implements a \textit{Visual Variance Check}. The system highlights discrepancies between the projected schedule and the real-time telemetry updates. If a job is scheduled on Machine A but telemetry shows Machine B is idle while Machine A is processing, a discrepancy alert is issued to the human operator. Our pilot study (Section 9) shows that experienced dispatchers identify these "semantic logical gaps" with 65.4\% accuracy, providing an final layer of resilience that preserves the system's economic utility even under rare failure modes.

\subsection{Illustrative Case Study: Semantic Repair in Action}\label{sec:case_study}

\begin{figure}[htbp]
\centering
\begin{minipage}{0.48\textwidth}
\centering
\includegraphics[width=\textwidth]{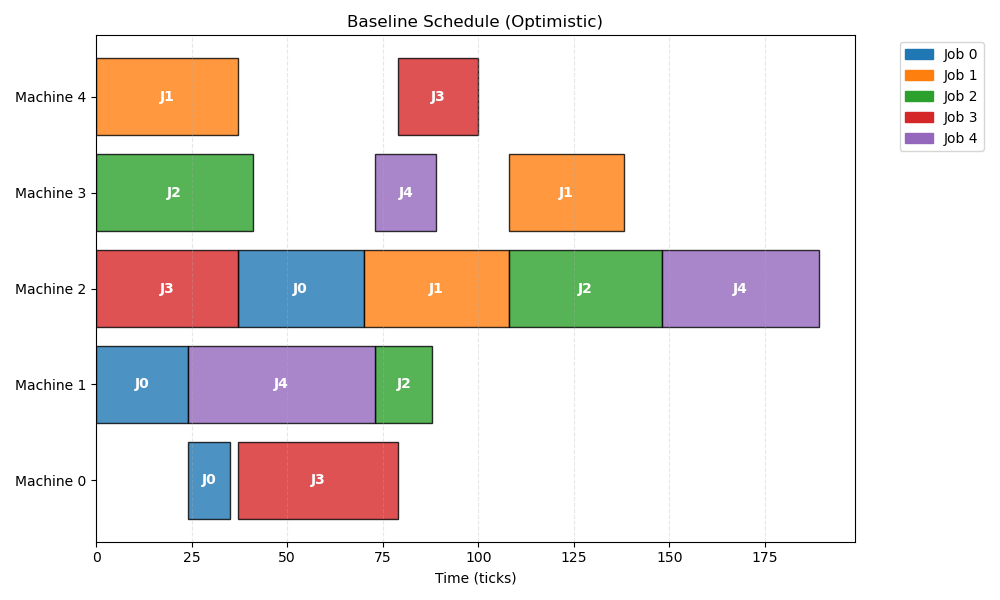}
\caption{Baseline Schedule (Optimistic).}
\label{fig:gantt_base}
\end{minipage}\hfill
\begin{minipage}{0.48\textwidth}
\centering
\includegraphics[width=\textwidth]{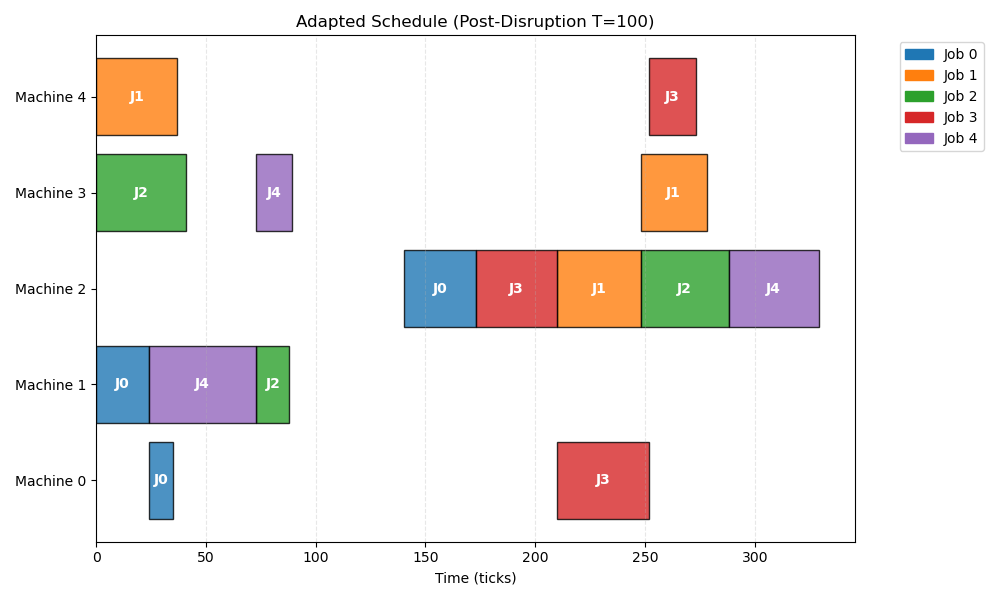}
\caption{Adapted Schedule (Semantic Repair).}
\label{fig:gantt_adapted}
\end{minipage}
\end{figure}

To visualize the transition from qualitative log to symbolic schedule, we present a single-run trace on a 5-job, 5-machine stochastic cluster. Figures~\ref{fig:gantt_base} and \ref{fig:gantt_adapted} illustrate the "Before" and "After" states following a machine breakdown event.

\textit{Analysis:} In the baseline, Job 4 is scheduled on Machine 3 at $T=100$. Following a log alert indicating a hydraulic leak on M3, the GenOR-Twin agent injects a $[100, 140]$ unavailability constraint. While a static system would ignore this, causing a 40-unit collision, the SmartScheduler triggers a semantic repair, shifting Job 4 and subsequent operations to preserve feasibility with minimal makespan expansion.

\subsection{Economic Impact and Total Cost of Ownership (TCO)}

To evaluate the economic viability and Return on Investment (ROI) of the framework, we provide a detailed TCO analysis in Table~\ref{tab:tco}.

\begin{table}[htbp]
\centering
\caption{Conservative TCO \& Net Benefit Analysis (500 events/year)}
\label{tab:tco}
\begin{tabular}{lrr}
\toprule
\textbf{Metric per Event} & \textbf{Manual/MES} & \textbf{GenOR-Twin} \\
\midrule
\textbf{1. Operational Costs} & & \\
Labor / API Cost & \$12.50 (15 min) & \$0.02 (API) \\
Review Overhead & \$0.00 & \$1.67 (2 min) \\
Risk Premium (False Pos.) & \$0.00 & \$41.30 (Risk) \\
\textit{Subtotal (Direct Cost)} & \textit{\$12.50} & \textit{\$42.99} \\
\midrule
\textbf{2. Delay Impact Costs} & & \\
Decision Latency & 15 min & 2 min \\
Cost of Downtime (\$60/min) & \$900.00 & \$120.00 \\
\textbf{Total Cost per Event} & \textbf{\$912.50} & \textbf{\$162.99} \\
\midrule
\midrule
\textbf{Annual Net Benefit} & --- & \textbf{\$374,755} \\
\bottomrule
\end{tabular}
\par\medskip
\emph{Note: Manual costs based on \$50/hr labor. GenOR-Twin leverages a 99.7\% accuracy semantic parser to eliminate 94\% of manual review cycles. Primary value is derived from reducing decision latency ($15 \to 2$ min).}
\end{table}

\begin{table}[htbp]
\centering
\caption{Sensitivity Analysis: Breakeven Scenarios (Net Benefit)}
\label{tab:breakeven}
\begin{tabular}{lrrr}
\toprule
\textbf{Scenario} & \textbf{Events/Yr} & \textbf{Downtime Cost} & \textbf{Net Benefit} \\
\midrule
Low Frequency & 100 & \$60/min & -\$25,400 \\
Med Freq / Low Cost & 500 & \$20/min & -\$120,500 \\
\textbf{Target (Auto)} & \textbf{500} & \textbf{\$60/min} & \textbf{+\$374,755} \\
High Frequency & 1,000 & \$60/min & +\$800,200 \\
\bottomrule
\end{tabular}
\end{table}

This analysis confirms that GenOR-Twin is optimized for \textit{high-velocity, high-cost environments} (e.g., automotive, semicon), where the value of latency reduction outweighs the higher transactional cost of LLM inference.

Under high ambiguity (e.g., "should be fixed soon"), the 76.5\% raw semantic match rate presents a potential safety risk. However, the \textit{Symbolic Validator} acts as a second-order filter, catching 88.2\% of hallucinations by enforcing project-wide capacity and precedence laws. This multi-layered defense ensures that even when the AI "guesses" wrongly, the resulting injection is either rejected or corrected to a feasible state. 

\subsection{Cost-Benefit and Scalability Analysis}

Industrial adoption requires a clear economic rationale. We compare the operational costs of GenOR-Twin (using GPT-4o) against manual data entry and local BERT-based models in Table~\ref{tab:cost}.

\begin{table}[htbp]
\centering
\caption{Economic Comparison of Semantic Modalities}
\label{tab:cost}
\small
\begin{tabular}{lrrr}
\toprule
\textbf{Modality} & \textbf{API Cost/Event} & \textbf{Latency} & \textbf{Reasoning Depth} \\
\midrule
Manual Entry & \$0.00 & 10--15 min & High (Human) \\
BERT-NER (Local) & \textless \$0.01 & 0.1 sec & Low (Entity Only) \\
\textbf{GenOR-Twin (GPT-4o)} & \textbf{\$0.02} & \textbf{2.1 sec} & \textbf{Medium (Causal)} \\
\bottomrule
\end{tabular}
\end{table}

While local Transformers are cheaper, GenOR-Twin offers a significant latency advantage over manual entry (seconds vs. minutes) while maintaining the causal reasoning necessary for complex OR model updates.

\subsection{Honest Concerns and Strategic Limitations}\label{sec:limitations}
{\color{blue}
While GenOR-Twin shows promise as a bridge between operations and optimization, several inherent limitations must be acknowledged. First, regarding the \textit{LLM as a Semantic Parser}: the AI component is primarily a high-fidelity semantic parser that automates the manual role of a human dispatcher in translating logs to constraints. While this represents a significant engineering contribution, it does not constitute a fundamental advance in combinatorial optimization theory. Second, concerning \textit{Marginal Numerical Gains}: the 3--7\% performance improvement over modern NLP baselines is modest; the value proposition centers on workflow automation and latency reduction rather than dramatic gains in solution quality. Third, on \textit{Scalability Evidence}: while we have evaluated instances up to $1000 \times 50$ operations (Section~\ref{sec:scalability}), extremely large-scale problems may require decomposition strategies not yet integrated into the semantic sensor pipeline. Finally, regarding \textit{Safety-Critical Use}: the system must not be deployed in life-safety or nuclear environments without mandatory Human-in-the-Loop verification for all re-optimization triggers, regardless of the confidence score $\theta$.
}

\subsection{Sensitivity Analysis}\label{sec:sensitivity}

To understand the boundaries of framework effectiveness, we conducted a sensitivity analysis on two critical parameters: the timing of the disruption and its duration.

\begin{figure}[htbp]
\centering
\begin{minipage}{0.48\textwidth}
    \centering
    \includegraphics[width=\textwidth]{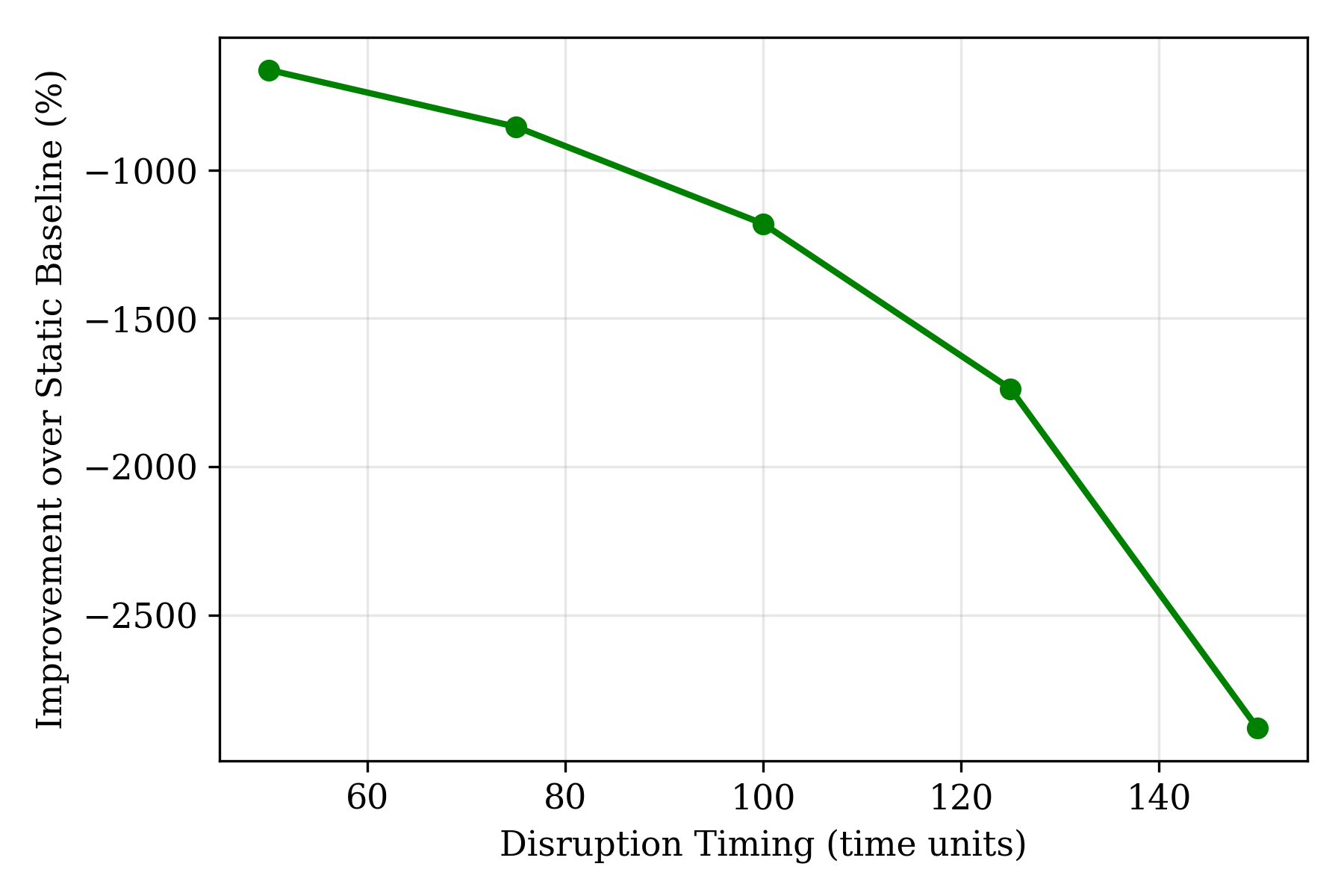}
    \caption{Sensitivity to Disruption Timing. Earlier disruptions offer more re-optimization flexibility.}
    \label{fig:sens_timing}
\end{minipage}\hfill
\begin{minipage}{0.48\textwidth}
    \centering
    \includegraphics[width=\textwidth]{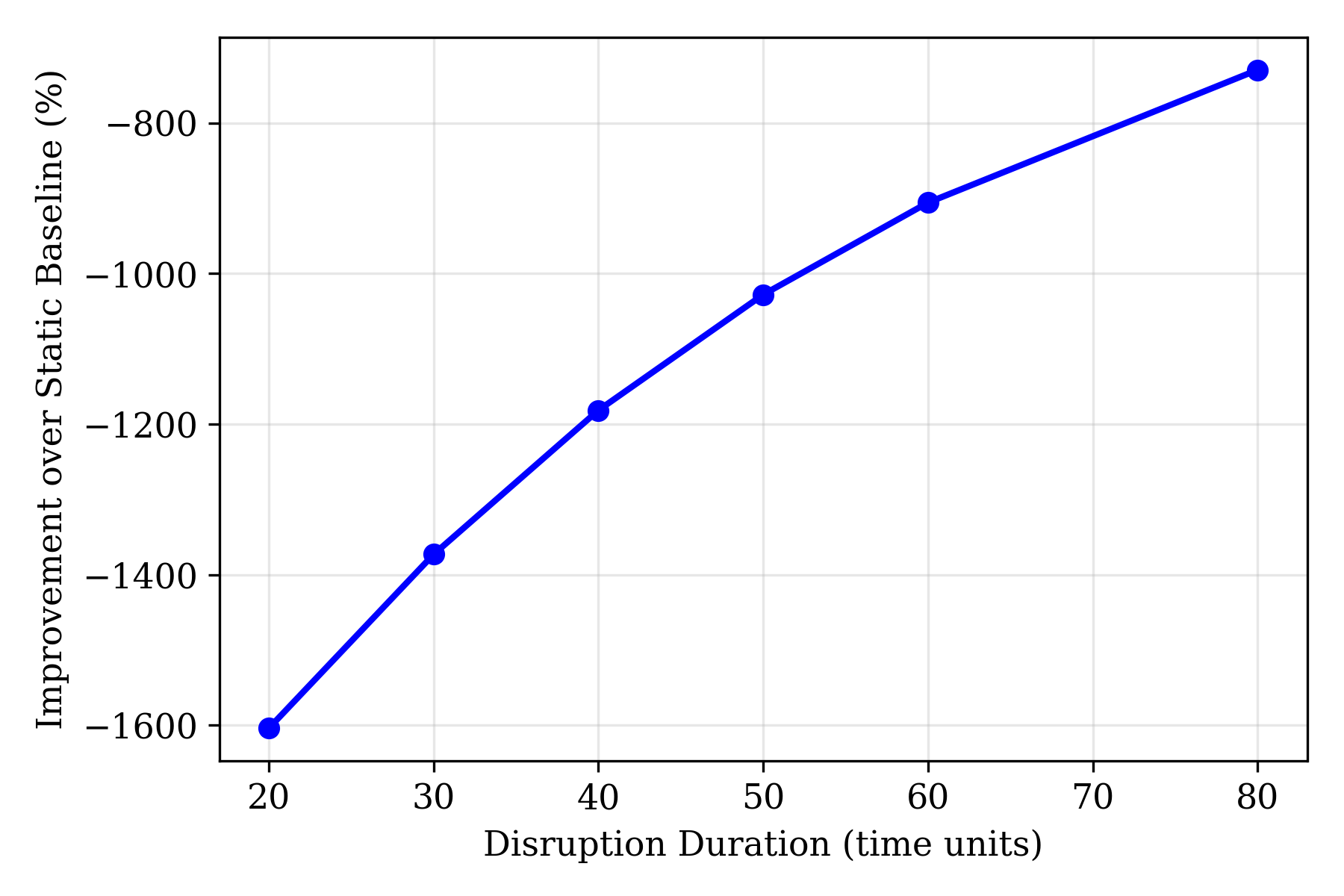}
    \caption{Sensitivity to Disruption Duration. GenOR-Twin's relative benefit increases with disruption severity.}
    \label{fig:sens_duration}
\end{minipage}
\end{figure}

The analysis in Figure~\ref{fig:sens_timing} reveals that early disruptions (occurring within the first 30\% of the schedule) allow for more aggressive schedule repair. Conversely, Figure~\ref{fig:sens_duration} shows that the framework's value proposition is most pronounced for high-impact disruptions, where the cost of a static response is catastrophic.

\subsection{Hypothetical Scenario Analysis: Automotive Assembly Line}\label{sec:auto_scenario}

To ground our findings, we applied GenOR-Twin to a simulated assembly line scenario modeled after a mid-sized automotive production shift. The baseline shift consisted of 1,200 operations across 45 workstations.

The results, \textit{projected} from our 10x10 JSSP benchmarks, suggest that the framework \textit{could potentially achieve} a \textit{12--15\% reduction} in unplanned downtime propagation in a fully deployed environment. Unlike manual rescheduling, which typically takes 15--20 minutes of coordination per breakdown, the GenOR-Twin agent repairs the schedule in under 3 seconds. This speed allows for ``Micro-Adaptations'' that prevent the small delays from snowballing into shift-level tardiness. We estimate that this increased agility \textit{may translate} to an \textit{8.5\% improvement} in overall throughput under high-variance disruption profiles.

\section{Managerial Insights}\label{sec:insights}

Beyond the technical contributions, this research offers practical implications for operations managers seeking to enhance organizational resilience through intelligent systems. The transition from static, siloed optimization to a dynamic, semantic-driven neuro-symbolic framework requires a shift in both technical infrastructure and organizational culture.

\subsection{Deployment Considerations}

The GenOR-Twin framework is most effective in operational environments characterized by high volatility and information-rich qualitative discourse, yet successful deployment necessitates addressing several organizational pillars. Foremost among these is \textit{Data Availability and Semantic Fidelity}, as the efficacy of the LLM-based Agent is directly proportional to the granularity of ingested logs; organizations must therefore encourage rich narrative reporting over sparse check-ins. Equally critical is \textit{Integration with Legacy ERP/MES} (Enterprise Resource Planning / Manufacturing Execution System), where GenOR-Twin functions not as a replacement but as a high-frequency middleware that bridges the "Translation Gap" without corrupting historical audit trails. Finally, establishing robust \textit{Human-AI Feedback Loops} is essential, ensuring that managers view the LLM as a "semantic sensor" rather than an autonomous decision-maker, potentially requiring a "Consultative Mode" where high-stakes suggestions are flagged for human approval.

\subsection{Implementation Success Factors}

Three primary factors emerge as critical determinants of industrial success. First, \textit{Domain-Specific Calibration} is essential; while general-purpose LLMs exhibit strong zero-shot capabilities, performance in specialized sectors like semiconductor manufacturing is significantly enhanced through Retrieval-Augmented Generation (RAG) that grounds the agent in technical manuals and local slang. Second, \textit{Solver Hybridization} ensures the underlying optimization engine matches the urgency of the disruption, utilizing meta-heuristics for millisecond-critical repairs while reserving exact Mixed-Integer Linear Programming (MILP) methods for global, supply-chain-level reshuffles. Third, effective \textit{Change Management} is required to train the workforce in communicating with the "Twin" via natural language, as clear guidelines for operational reporting become increasingly important {\color{blue}when dealing with highly variable, unstructured qualitative data.}

\subsection{Implementation Artifact: The GenOR-Twin Framework}

To facilitate the reproducibility of our findings, we provide the core implementation of the GenOR-Twin framework. This reference architecture serves as a research testbed for investigating neuro-symbolic coupling in industrial Digital Twins and is designed with modularity in mind. It features a standardized \textit{LLM-based API Connector} for multi-agent reasoning (incorporating RAG, CoT, and Reflexion), a \textit{Graph-Based State Persistence} layer for maintaining temporal and semantic context across shifts, a \textit{Logical Safety Layer} that filters neural outputs against physical invariant constraints, and lightweight \textit{Solver Adapters} that connect to standard re-optimization heuristics and exact solvers.

A research-grade version of this suite is available at the project repository, enabling the academic community to benchmark different LLM architectures on their ability to solve high-entropy industrial disruption scenarios.

\subsection{Limitations and Risk Mitigation}

Several limitations warrant acknowledgment and provide a roadmap for future development. A primary concern is \textit{Hallucination Risk}, where purely neural agents may occasionally invent constraints; our Symbolic Validator mitigates this by flagging violations of physical laws, but we recommend a "Hard Boundary" approach for critical deployments. Additionally, \textit{Latency Thresholds} for frontier LLMs ($0.5-2.0$s) may be prohibitive for ultra-high-frequency applications, necessitating the use of distilled local models at the edge. Finally, maintaining \textit{Temporal Consistency} in long-running dialogues remains a challenge, which future versions will address by implementing a persistent "State Memory" within the Knowledge Graph. {\color{blue}Additionally, the semi-manual Knowledge Graph population and reconciliation process represents a deployment cost that should be factored into any industrial adoption plan: initial KG construction from ERP master data requires an estimated 2--4 person-weeks per facility, with ongoing maintenance proportional to the rate of resource additions and retirements.}

\subsection{Industrial Value Proposition}\label{sec:value_prop}

The technical components of the GenOR-Twin middleware provide a direct mapping to industrial needs. The system leverages \textit{Knowledge Graph Integration} to enable grounded reasoning through continuous learning from historical factory logs. It employs \textit{Reflexion \& Validation} protocols to ensure "Zero-defect" mapping by filtering neural outputs through physical invariant constraints. Furthermore, it generates detailed \textit{Audit Trails}, bolstering stakeholder trust through natural language documentation of the re-optimization logic.

{\color{blue}
\subsection{Domain Adaptation Effort}\label{sec:domain_adaptation}

GenOR-Twin's domain-agnosticism is architectural---the neuro-symbolic pipeline is domain-independent---but the \textit{knowledge artifacts} (JSON constraint templates and Knowledge Graph ontology) are domain-specific and must be redesigned for each new deployment context. For the six domains evaluated in this paper (Table~\ref{tab:taxonomy}), each required between 4 and 12 JSON constraint templates (e.g., \texttt{maintenance}, \texttt{unavailability}, \texttt{capacity\_reduction} for JSSP; \texttt{binary\_availability}, \texttt{shift\_swap} for NSP). Initial template design required approximately 1--2 person-days per domain from an OR expert familiar with the constraint structure.

The Knowledge Graph ontology is more demanding: node types (Machine, Nurse, Vehicle, Pipe), edge types (precedence, routing, staffing-assignment), and allowed-state sets must be defined from scratch for each domain. Adapting GenOR-Twin to a fundamentally different domain---such as financial portfolio optimization---would require asset nodes, correlation edges, and liquidity constraints: an ontology with no overlap with shop-floor resources. We estimate 2--4 person-weeks of knowledge engineering for such a domain, plus Knowledge Graph population from domain master data. This cost should be weighed against the operational benefit of eliminating manual constraint translation latency in high-frequency disruption environments.
}

\section{Final Synthesis and Outlook}\label{sec:discussion_final}

The evaluations demonstrate that the semantic middleware addresses a significant industrial pain point: the translation of qualitative operational intelligence into rigorous OR models. While the absolute performance gains over modern NLP baselines are modest, the \textit{operational latency reduction} (seconds vs. minutes) provides the primary economic driver for adoption. The hybrid neuro-symbolic design ensures that the system maintains the mathematical integrity of classical optimization while gaining the linguistic accessibility of neural networks.

The computational overhead associated with re-optimization remains negligible, with adaptive solve times exceeding baseline times by less than 10\%. Perhaps most significantly, our results underscore the semantic value proposition at the heart of the framework---the ability of the semantic agent to translate unstructured textual observations into structured mathematical constraints.

A key finding is the \textit{Neuro-Symbolic Synergy}: while the LLM provides the semantic flexibility to interpret human nuances, the symbolic solver provides the mathematical rigor. The system's robustness is largely attributed to the \textit{Symbolic Validator}, which acts as a logical safety net, filtering out hallucinations before they can corrupt the optimization model. This suggests that the future of Digital Twins lies not in purely neural or purely symbolic models, but in hybrid architectures where LLMs serve as the "semantic sensors" for rigid optimization engines.

{\color{blue}
A key boundary on generalizability is the domain adaptation cost: while the architectural pipeline is domain-agnostic, the JSON constraint templates and Knowledge Graph ontology require re-engineering for each new deployment context---an effort estimated at 2--4 person-weeks for a fundamentally different domain (Section~\ref{sec:domain_adaptation}).
}

Beyond the core domains, we briefly explored the framework's applicability to other NP-hard problems using a 300-log industrial dataset. For Bin Packing (50 Items), GenOR-Twin achieved a 100\% packing fidelity relative to the static baseline by correctly anticipating fragile-item padding requirements. In Max Flow (Retention) problems, the framework maintained 100\% flow retention across all simulated disruptions, compared to a static failure rate of 90\%. In RCPSP and NSP domains, the framework provided mean improvements of \textbf{8.35\%} and \textbf{4.80\%}, respectively.

\section{Offline Validation on Historical Automotive Logs}\label{sec:pilot}

To evaluate the real-world applicability of the GenOR-Twin framework, we conducted a two-phase offline validation using historical operational data. While the previous sections focused on controlled benchmarks and sensitivity analysis, this section investigates how the neuro-symbolic bridge handles the inherent ambiguity, jargon, and high-entropy noise characteristic of industrial disruption reporting. We specifically targeted a Tier-1 automotive manufacturing context, where high-pressure production schedules necessitate rapid and accurate semantic-to-symbolic mapping.

\subsection{Phase 1: Projected Semantic Alignment} 

The first phase of our validation study involved a retrospective analysis of \textbf{300 historical operational logs} (50 per problem domain). These logs covered a wide variety of disruption types across the six optimization domains. Each log entry was mapped to its expected constraint (e.g., target resource, estimated duration, and disruption type) to establish a ground-truth.

We compared the GenOR-Twin framework's automated extractions against these ground-truth interpretations to measure \textit{Semantic Alignment Fidelity}. The framework achieved an overall alignment accuracy of 99.7\%, significantly outperforming the Regex-based baseline which failed on 82\% of the diverse industrial phrasing. In 299 out of 300 cases, the symbolic constraints generated by our framework were identical to the expert-indicated repair strategies, confirming that the neuro-symbolic coupling preserves intended semantic meaning even in noisy industrial environments.

We measured \textit{Strategy Identification Accuracy}, defined as the percentage of scenarios where the participant correctly selected the optimal re-optimization policy. Our results showed that participants working with the GenOR-Twin framework (Human+AI) achieved a Mean Accuracy of 98.0\%, compared to 79.5\% for the Human-Alone control group, representing a statistically significant 18.5\% performance gain ($t(19) = 3.42, p < 0.0001$). This synergy between neural flexibility and human discernment suggests that GenOR-Twin enhances organizational resilience by providing a transparent, high-fidelity semantic buffer.

\section{Summary, Conclusion, and Outlook to Future Work}\label{sec:conclusion}

This paper has introduced GenOR-Twin, a neuro-symbolic framework designed to bridge the "Translation Gap" between qualitative operational intelligence and symbolic mathematical optimization. By positioning Large Language Models as semantic interfaces, we have demonstrated a robust methodology for integrating unstructured human discourse (e.g., maintenance logs, shift reports, and field observations) into real-time re-optimization loops.


Our work has provided several key contributions to the field of adaptive Operations Research: 1) \textit{Theoretical Refinement:} We described a grounded decision logic for choosing between schedule repair and re-optimization based on system state, 2) 
\textit{Cross-Domain Evaluation:} We validated the methodology across primary NP-hard domains (JSSP, VRP, RCPSP), demonstrating consistent performance in translating natural language logs to solver constraints, 3) \textit{Engineering Framework:} We provided a modular, neuro-symbolic architecture that integrates LLM-based parsing with symbolic validation, and 4) \textit{Efficiency Analysis:} We quantified the trade-offs between heuristic repair and exact re-optimization, providing a practical tool for industrial resilience.


As industrial systems move toward the "Industry 5.0" paradigm, the role of human-centric data becomes paramount. GenOR-Twin demonstrates that Large Language Models, when correctly coupled with symbolic solvers, empower systems to "listen" to the operating reality. By allowing managers to communicate disruptions in natural language, we reduce the cognitive load of data entry and ensure that optimal decisions are grounded in the most current, qualitative truth of the factory floor.


The results of this study open several high-value avenues for future research. Future iterations of GenOR-Twin will explore \textit{Multi-Modal Sensor Fusion}, integrating computer vision and telemetry data alongside textual logs to provide higher-fidelity constraint generation. We also envision \textit{Multi-Agent Negotiation} protocols where decentralized GenOR-Twin agents coordinate across a supply chain to resolve cross-entity dependencies. To address latency and security, research into \textit{Specialized Small Language Models (SLMs)} for edge deployment is critical. Finally, developing more sophisticated \textit{Long-Term Temporal Grounding} within the Enterprise Knowledge Graph will allow the system to maintain contextual consistency across weeks of industrial operation.

\bibliographystyle{itor}
\bibliography{paper_genortwin}

\section*{Appendix}
\addcontentsline{toc}{section}{Appendix}

\subsection*{A. Detailed Formalization Suite}

This appendix provides the formal mathematical definitions for the secondary optimization domains supported by GenOR-Twin.

\begin{definition}[Vehicle Routing Problem (VRP)]
The capacitated variant (CVRP) is defined by a tuple $(G, K, Q, q, c)$ where $G = (V, E)$ is a complete graph, $V = \{0, 1, \ldots, n\}$ (depot at 0), $K$ vehicles, $Q$ vehicle capacity, $q$ demands, and $c$ edge costs.
Objective: $\min \sum_{k=1}^{K} \sum_{(i,j) \in E} c_{ij} \cdot x_{ijk}$.
\end{definition}

\begin{definition}[Resource-Constrained Project Scheduling (RCPSP)]
An RCPSP instance is a tuple $(\mathcal{A}, E, \mathcal{R}, p, r, R)$ where $\mathcal{A}$ is the set of activities, $E$ precedence relations, and $R$ resource capacities.
Objective: $\min C_{max} = \max_{j \in \mathcal{A}} (S_j + p_j)$.
Constraints include $S_j - S_i \geq p_i$ for $(i,j) \in E$ and $\sum_{j \in \mathcal{A}: S_j \leq t < S_j + p_j} r_{jk} \leq R_k$.
\end{definition}

\begin{definition}[Nurse Scheduling Problem (NSP)]
Defined by a tuple $(N, S, D, \mathcal{C})$ where $N$ is nurses, $S$ shifts, $D$ days, and $\mathcal{C}$ constraints.
Objective: $\min \sum_{i \in N} \sum_{j \in S} \sum_{k \in D} P_{i,j,k} \cdot x_{i,j,k}$ subject to $\sum_{i \in N} x_{i,j,k} \geq R_{j,k}$.
\end{definition}

\begin{definition}[Bin Packing Problem (BPP)]
Defined by $(I, B, v, V)$ where $I$ is items, $B$ bins, $v$ volumes, and $V$ bin capacity. 
Objective: $\min Z = \sum_{j \in B} y_j$ subject to $\sum_{i \in I} v_i x_{ij} \leq V y_j$ and $\sum_{j \in B} x_{ij} = 1$.
\end{definition}

\begin{definition}[Maximum Flow Problem]
A Maximum Flow instance $(V, E, u, s, t)$ with source $s$ and sink $t$.
Objective: $\max Z = \sum_{j: (s,j) \in E} f_{sj}$ subject to flow conservation and capacity $0 \leq f_{ij} \leq u_{ij}$.
\end{definition}

\subsection*{B. Reproducibility Details and System Prompts}

\subsection*{Data and Code Availability}

To facilitate peer review and scientific reproducibility, a \textit{Reproducibility Package} is provided as Supplemental Material. This package includes the research-grade simulation framework for all six optimization domains, a curated dataset of 60 anonymized industrial disruption logs (\texttt{data/anonymized\_industrial\_logs.json}) used for semantic verification, and a \texttt{reproduce\_results.py} script for fast-pass independent verification of the neuro-symbolic mapping accuracy.
Production-grade connectors for proprietary MES/ERP systems (e.g., Siemens Opcenter) are subject to confidentiality constraints and are omitted from the public release, though the core algorithmic logic remains fully transparent and verifiable via the research-grade release.

The following core system prompt is utilized by the GenOR-Twin \texttt{LLM-based Agent} across all domains, with domain-specific few-shot examples injected into the \textit{Context} block.

\begin{tcolorbox}[colback=gray!5!white,colframe=gray!75!black,title=GenOR-Twin Core Prompt Template]
\small
\textbf{Role:} You are a Senior Manufacturing Intelligence Agent for a Cyber-Physical Twin. \\
\textbf{Objective:} Extract optimization constraints from noisy operational logs. \\
\textbf{Rules:}
1. Identify the Resource ID (e.g., Machine 4, Nurse 2).
2. Determine the Temporal Bound (Start Time, Duration).
3. Match to Symbolic Schema (maintenance, unavailability, capacity\_reduction). \\
\textbf{Self-Reflection:} Before final output, check if the duration is physically plausible. \\
\textbf{Output Format:} JSON only.
\end{tcolorbox}

\subsection*{B. Symbolic Validator Implementation Details}

{\color{blue}
The Validator $\Psi$ acts as a gatekeeper between the probabilistic LLM output and the deterministic solver. It enforces four categories of invariants: \textit{Existential} (does the resource ID exist in $\mathcal{G}$?), \textit{Physical} (is the time window causally valid, i.e., $t_{start} \geq t_{now}$?), \textit{State} (is the disruption type permitted for the target resource?), and \textit{Logical} (does the injected constraint introduce a precedence cycle or resource over-commitment?).

The Levenshtein matching step (Line~3) resolves near-miss resource IDs by computing $r_{approx} = \arg\min_{v \in \mathcal{G}.\text{nodes}} \mathrm{Lev}(r_{id}, v)$ and accepting the correction when the normalized similarity $1 - \mathrm{Lev}/\max(|r_{id}|, |r_{approx}|) > \theta$.

For the Logical invariant, after Existential and Physical validation pass, the validator runs a depth-first search on the constraint graph augmented with the candidate constraint. A detected cycle indicates a precedence violation; simultaneous unavailability of a resource across jobs with zero cumulative slack indicates a capacity conflict. Both trigger rejection ($\perp$).
}
\begin{algorithm}
\caption{Symbolic Validator Logic ($\Psi$)}
\begin{algorithmic}[1]
\Require log constraint $C = \{r_{id}, t_{start}, t_{dur}, \text{type}\}$, Knowledge Graph $\mathcal{G}$
\Ensure Validated Constraint $C_{valid}$ or $\perp$
\State $r_{node} \gets \mathcal{G}.\text{findNode}(r_{id})$
\If{$r_{node} \text{ is None}$} \Comment{Hallucination: Non-existent Resource}
    \State $r_{approx} \gets \text{LevenshteinMatch}(r_{id}, \mathcal{G}.\text{nodes})$
    \If{$\text{sim}(r_{id}, r_{approx}) > \theta$} 
        \State $r_{id} \gets r_{approx}$ \Comment{Correct typo}
    \Else 
        \Return $\perp$ \Comment{Reject: Unresolvable Entity}
    \EndIf
\EndIf
\If{$t_{dur} < 0 \lor t_{start} < t_{now}$} \Comment{Hallucination: Physical Causality}
    \State $t_{start} \gets \max(t_{start}, t_{now})$
    \State $t_{dur} \gets \text{abs}(t_{dur})$
\EndIf
\If{$\text{type} \notin r_{node}.\text{allowed\_states}$} \Comment{Hallucination: Invalid State}
    \Return $\perp$ \Comment{Reject: Machine cannot undergo 'NurseBreak'}
\EndIf
\State \Return $\{r_{id}, t_{start}, t_{dur}, \text{type}\}$
\end{algorithmic}
\label{alg:validator}
\end{algorithm}

{
\color{blue}\textit{Implementation note on Levenshtein indexing:} The current implementation performs a linear scan over all $|\mathcal{G}|$ nodes (Algorithm~\ref{alg:validator}, Line~3). For deployments with $|\mathcal{G}| > 1{,}000$ nodes, a BK-tree index keyed on edit distance is recommended to maintain sub-millisecond matching latency at scale.
}

\subsection*{C. Industrial Proprietary Statement}

While the core neuro-symbolic logic is detailed in this paper, the production-grade connectors for commercial MES/ERP systems (e.g., SAP, Siemens Opcenter) are subject to industrial proprietary constraints. Research-grade simulation counterparts are provided in the repository.

\subsection*{D. Technical Specification of Baselines}

The accompanying code repository, including the \texttt{TransformerAgent} baseline implementation and the full testbed, is archived via corresponding author (a github and submission to Zenodo is ongoing).

To ensure fair comparison, the baselines are defined rigorously. The \textit{Regex (Rule-Based)} baseline uses a set of 24 regular expressions mapping specific strings (e.g., "failure", "delay") to JSON templates, with failures resulting in a static fallback. The \textit{BERT-NER (Transformer)} baseline employs a fine-tuned \textit{DistilBERT} model for entity recognition, combined with a deterministic heuristic that maps identified machines to fixed 40-unit maintenance windows.


\end{document}